\documentclass[twoside,11pt]{article}
\usepackage{microtype}

\usepackage{jmlr2e}

\usepackage{amsmath}
\usepackage{bbm} 
\usepackage{booktabs}

\usepackage{xargs}  
\usepackage[pdftex,dvipsnames]{xcolor}  

\newcommand{\R}{\mathbb{R}}

\newcommand{\Pm}{\textnormal{P}}
\newcommand{\x}{\mathbf{x}}

\newcommand{\Ts}{T^*}
\newcommand{\kat}{{\kappa(t)}}
\newcommand{\kati}{{\kappa(t_i)}}

\renewcommand{\mid}{\,|\,}
\newcommand{\interv}[2]{\mathopen{(} #1,\,  #2 \mathclose{]}}
\newcommand{\intervl}[2]{\mathopen{[} #1,\,  #2 \mathclose{)}}

\ShortHeadings{Continuous and Discrete-Time Survival Prediction with Neural Networks}{Kvamme, Borgan}
\firstpageno{1}

\begin{document}

\title{Continuous and Discrete-Time Survival Prediction\\with Neural Networks}

\author{\name{H\aa{}vard Kvamme} \email{haavakva@math.uio.no} \\
        \name{\O{}rnulf Borgan} \email{borgan@math.uio.no} \\
        \addr{Department of Mathematics\\
        University of Oslo\\
        P.O. Box 1053 Blindern\\
        0316 Oslo, Norway}
        }


\maketitle

\begin{abstract}
Application of discrete-time survival methods for continuous-time survival prediction is considered.
For this purpose, a scheme for discretization of continuous-time data is proposed by considering the quantiles of the estimated event-time distribution, and, for smaller data sets, it is found to be preferable over the commonly used equidistant scheme.
Furthermore, two interpolation schemes for continuous-time survival estimates are explored, both of which are shown to yield improved performance compared to the discrete-time estimates.
The survival methods considered are based on the likelihood for right-censored survival data, and parameterize either the probability mass function (PMF) or the discrete-time hazard rate, both with neural networks.
Through simulations and study of real-world data, the hazard rate parametrization is found to perform slightly better than the parametrization of the PMF\@.
Inspired by these investigations, a continuous-time method is proposed by assuming that the continuous-time hazard rate is piecewise constant. The method, named PC-Hazard, is found to be highly competitive with the aforementioned methods in addition to other methods for survival prediction found in the literature.
\end{abstract}

\begin{keywords}
    survival analysis, neural networks, time-to-event prediction, interpolation, discretization
\end{keywords}

\section{Introduction}

Survival analysis, or time-to-event analysis, considers the problem of modeling the time of a future event. A plethora of statistical methods for analyzing right-censored time-to-event data have been developed over the last fifty years or so. Most of these methods, like Cox regression, assume continuous-time models,  but methods based on discrete-time models are sometimes used as well. For a review see, e.g., \citet{klein2005survival}  for statistical methods based on continuous-time models  and \citet{tutz2016} for discrete-time models and methods.

As a result of the rapid development in machine learning and, in particular, neural networks, a number of new methods for time-to-event predictions have been developed in the last few years. This development has benefited from the excellent frameworks for neural network development, such as TensorFlow, PyTorch, Theano, Keras, and CNTK, which have simplified the application of neural networks to existing likelihood-based methodology.
Thus, novel methods for time-to-event predictions have been developed based on Cox partial likelihood~\citep[e.g.,][]{DeepSurv, Luck2017, yousefi2017predicting, Cox-Time} and the discrete-time survival likelihood~\citep[e.g.,][]{deephit, fotso2018, gensheimer2019}.

To the best of our knowledge, \citet{deephit} were the first to apply neural networks to the discrete-time likelihood for right-censored time-to-event data. Their approach was to parameterize the probability mass function (PMF) of the event times.
In statistical survival analysis, it is, however, more common to  express the likelihood by the discrete-time hazard rate~\citep[see, e.g.,][]{tutz2016}. \citet{gensheimer2019} used this form of the  likelihood and parameterized the hazard rates with a neural network. In this paper we perform a systematic study of the use of neural nets in conjunction with  discrete-time  likelihoods for right-censored time-to-event data; in particular, we perform a comparison of methods that parameterize the PMF and the discrete hazard rate.

It is common to apply discrete-time methods as an approximation for continuous-time survival data.
To this end  one has to perform a  discretization of the continuous time scale; a subject that has received little attention in the literature.
We consider two discretization schemes, corresponding to equidistant times or equidistant survival probabilities, and conduct a simulation study to better understand the effect of the discretization scheme and the number of time-points used for the discrete-time methods.

Closely related to the discretization of a continuous time scale is the subject of interpolation.
A coarse discretization grid has the benefit of reducing the number of parameters in a neural network. But the approximation error that incurs  when a discrete-time method is used as an approximation for  continuous-time data, becomes smaller with a denser grid.  By interpolating the discrete-time survival estimates, it is possible to use a coarser discretization grid without increasing the approximation error.
For this reason, two simple interpolation schemes are investigated in this paper; the first assumes constant density functions between the time-points in the discretization grid, and the second assumes constant hazard rates between the grid points. As a modification of the latter method, we also
propose a continuous-time method obtained by assuming that the continuous-time hazard rate is piecewise constant, and we compare this method with the aforementioned discrete-time methods with and without interpolation.

The paper is organized as follows.
First, in Section~\ref{sec:related_works}, we present a summary of related methods for time-to-event predictions. Then,
in Section~\ref{sec:discrete_time_models}, we consider the discrete-time likelihood for right-censored event-times and discuss how the likelihood may be parameterized with neural networks.
In Section~\ref{sec:continuous_time_models}, continuous-time models for time-to-event data are considered, and
we discuss how discretization of the continuous time scale enables the application of discrete-time survival methods for continuous-time data. Here we also present the two schemes for interpolating  discrete survival functions, and we consider our continuous-time method obtained by assuming piecewise constant hazards.
In Section~\ref{sec:simulations}, a simulation study is conducted to understand the impact the discretization and interpolation schemes have on the methods, and in Section~\ref{sec:experiments}, we compare the methods with existing methods for time-to-event predictions using five real-world data sets.
Finally, we summarize and discuss our findings in Section~\ref{sec:discussion}.
The code for all methods, data sets, and simulations presented in this paper is available at \url{https://github.com/havakv/pycox}.

\section{Related Works}
\label{sec:related_works}

Numerous researchers have used neural networks to parameterize the likelihood for discrete-time survival data.
If fact, the two discrete-time survival methods explored in this paper were first proposed by \citet{deephit} and \citet{gensheimer2019}.

DeepHit~\citep{deephit} parameterizes the probability mass function  (PMF) of the survival distribution and combines the log-likelihood for right-censored data with a ranking loss for improved discriminative performance. The method has been extended to allow  for competing risks data.
 \citet{deephit} only used the time-dependent concordance of \citet{Antolini2005} for performance evaluation, and they did not discuss discretization of a continuous time scale.
\citet{Cox-Time} showed that, by only considering concordance, DeepHit has excellent discriminative performance at the cost of poorly calibrated survival estimates.

It is well known in the survival analysis literature that the log-likelihood of discrete-time survival data can be expressed as a Bernoulli log-likelihood of the  hazard rates.
This enables the use of generalized linear models (GLM) software for fitting survival models parameterized by the hazard rate; for an overview see \citet{tutz2016}.
\citet{gensheimer2019} extended this methodology by parameterizing the hazard rates with a neural network.
They showed that their method performs well, both in terms of discrimination and calibration of the survival predictions. However, they did not compare their methodology with methods that parameterize the PMF\@.

\citet{MTLR} proposed the multi-task logistic regression, which is a generalization of the binomial log-likelihood, to jointly model a sequence of binary labels representing event indicators.
\citet{fotso2018} later applied this framework to neural networks.
We show in Section~\ref{sub:multi_task_logistic_regression} that the multi-task logistic regression is, in fact, a PMF parametrization.

Another approach to time-to-event prediction is to consider time as continuous rather than discrete. As a result, the obtained methodology is often not fully parametric.
Many of the proposed continuous-time methods are based on Cox proportional hazards model, also called Cox regression model. Estimation in this semi-parametric model is commonly based on Cox partial likelihood~\citep[see, e.g.,][]{klein2005survival}.
\citet{Faraggi1995} were the first to parameterize a Cox regression model with a neural network. They were, however, unsuccessful in achieving any improvements over regular Cox regression.

Later extensions of the Cox proportional hazards methodology include new network architectures, larger data sets, and better optimization schemes \citep{DeepSurv, ching2018cox, yousefi2017predicting}.
As a result, the predictive performance has been improved, in addition to enabling covariates in the form of images \citep{7822579, 8100208}.
\citet{Luck2017} combined the negative Cox partial log-likelihood with an isotonic regression loss in an attempt to obtain better discriminative performance. Regardless, their method is still limited by the  proportional hazards assumption.

The proportionality assumption of Cox regression model is quite restrictive. Unlike the discrete methods discussed above, none of the aforementioned Cox extensions can estimate survival curves that cross each other.
\citet{Cox-Time} alleviated this restriction by proposing a non-proportional extension of the Cox methodology.
This was achieved by approximating the partial log-likelihood with a loss based on case-control sampling.

Random Survival Forest (RSF) by~\cite{Ishwaran2008} is a fully non-parametric continuous-time method for right-censored survival data.
RSF computes decision trees based on the log-rank test and estimates the cumulative hazard rate with the Nelson-Aalen estimator.
The RSF method has become a staple in the predictive survival literature, and it is used as a benchmark in the majority of the work listed in this section.

\section{Discrete-Time Models}
\label{sec:discrete_time_models}

In this section, we will restrict ourselves to models in discrete time.
Then, in Section~\ref{sec:continuous_time_models}, we will discuss how discrete-time models may be used as approximations of models in continuous time.
In the following, we start by a brief introduction to terms in the field of survival analysis, followed by the derivation of the likelihood for right-censored survival data, which is the basis for all methods presented in this paper (and much of survival analysis in general).
We will then show how we can parameterize the likelihood with neural networks to obtain the methods proposed by 
\citet{deephit}, \citet{gensheimer2019}, \citet{MTLR}, and \citet{fotso2018}.

\subsection{The Discrete-Time Survival Likelihood}
\label{sub:the_discrete_survival_likelihood}

Assume that time is discrete with values $0 = \tau_0 < \tau_1 < \ldots$, and let $\mathbb T = \{\tau_1, \tau_2, \dots \}$ denote the set of positive $\tau_j$'s.
The time of an event is denoted $T^* \in \mathbb T$, and our goal is to model the distribution of such event times, or durations.
The probability mass function (PMF) and the survival function for the event times are defined as
\begin{align}
    \label{eq:survival_orig}
    f(\tau_j) &= \Pm(T^* = \tau_j),\nonumber\\
    S(\tau_j) &= \Pm(T^* > \tau_j) = \sum_{k > j} f(\tau_k).
\end{align}
In survival analysis, models are often expressed in terms of the hazard function rather than the PMF\@.
For discrete time, the hazard is defined as 
\begin{align*}
    h(\tau_j) &= \Pm(T^* = \tau_j \mid T^* > \tau_{j-1}) = \frac{f(\tau_j)}{S(\tau_{j-1})} = \frac{S(\tau_{j-1}) - S(\tau_j)}{S(\tau_{j-1})},
\end{align*}
and it follows that
\begin{align}
    \label{eq:hazard_more_f}
    f(\tau_j) &= h(\tau_j)\, S(\tau_{j-1}),\\
    \label{eq:hazard_more_S}
    S(\tau_j) &= [1 - h(\tau_j)]\, S(\tau_{j-1}).
\end{align}
Note further, that from~\eqref{eq:hazard_more_S} it follows that the survival function can be expressed as
\begin{align}
    \label{eq:surv_hazard}
    S(\tau_j) &= \prod_{k=1}^j [1 - h(\tau_k)].
\end{align}

In most studies, we do not observe all event times. For some individuals, we will only observe a right-censored duration.
So to allow for censoring, we let $C^* \in \mathbb T_C = \{\tau_1, \tau_2, \ldots, \tau_m\}$ be a right-censoring time.
In the same manner as for the event time, the censoring-time has the PMF and survival function
\begin{align*}
    f_{C^*}(\tau_j) &= \Pm(C^* = \tau_j), \\
    S_{C^*}(\tau_j) &= \Pm(C^* > \tau_j).
\end{align*}
$T^*$ and $C^*$ are typically not observed directly, but instead, we observe a potentially right-censored duration $T$ and an event indicator $D$ given by
\begin{align*}
    &T = \min\{T^*,\, C^*\},\\
    &D = \mathbbm{1}\{T^* \leq C^*\}.
\end{align*}
We here follow the common convention in survival analysis that when an event and censoring time coincide, we observe the occurrence of the event. 
Note that, as $C^* \leq \tau_m$, we are not able to observe event times $T^*$ larger than $\tau_m$. Hence, we are restricted to model the distribution of the event times in $\mathbb T_C$.

Now, assuming that $\Ts$ and $C^*$ are independent, we can derive the likelihood function for right-censored survival data.
To this end, note that, for $t \in \mathbb{T_C}$ and $d \in \{0, 1\}$, we have that
\begin{align*}
    \Pm(T = t, D=d) \nonumber
    &= \Pm{(T^*=t,\, C^* \geq t)}^d \; {\Pm(T^* > t,\, C^*=t)}^{1-d} \nonumber \\
    &= {\left[\Pm(T^*=t)\, \Pm(C^* \geq t)\right]}^d \; 
    {\left[\Pm(T^* > t)\, \Pm(C^*=t)\right]}^{1-d} \nonumber  \\
    &= {\left[f(t)\, (S_{C^*}(t) + f_{C^*}(t)) \right]}^d \,  {\left[S(t) f_{C^*}(t) \right]}^{1-d} \nonumber\\
    &= \left[{f(t)}^d \,  {S(t)}^{1-d}\right] \left[ {f_{C^*}(t)}^{1-d}\, {(S_{C^*}(t) + f_{C^*}(t))}^d\right].
\end{align*}
Now, it is common to assume that $f(t)$ and $f_{C^*}(t)$ have no parameters in common.
Then we can consider, separately, the contribution to the likelihood of the event time distribution and the censoring distribution.
We are typically only interested in modeling the distribution of the event times, in which case, for individual $i$, we obtain the likelihood contribution
\begin{align}
    \label{eq:likelihood_survival}
    L_i &= {f(t_i)}^{d_i} {S(t_i)}^{1-d_i}.
\end{align}
If we have data for $n$ independent individuals, each with covariates $\x_i$, observed time $t_i$, and event indicator $d_i$, we can fit models by minimizing the mean negative log-likelihood
\begin{align}
    \label{eq:loss_general}
    \text{loss} &= - \frac{1}{n} \sum_{i=1}^n \left( d_i \log [f(t_i \mid \x_i)] + (1-d_i) \log [S(t_i \mid \x_i)] \right).
\end{align}

A useful alternative to the loss function~\eqref{eq:loss_general} is obtained by rewriting it in terms of the discrete hazards.
To this end, let $\kappa(t) \in \{0, \ldots, m\}$ define the index of the discrete time $t$, meaning $t = \tau_\kat$.
Using~\eqref{eq:hazard_more_f},~\eqref{eq:hazard_more_S}, and~\eqref{eq:surv_hazard}, we can then rewrite the likelihood contribution~\eqref{eq:likelihood_survival} as
\begin{align*}
    L_i &= {f(t_i)}^{d_i} \, {S(t_i)}^{1-d_i} \\
    &= {[h(t_i)\, S(\tau_{\kati-1})]}^{d_i} \, {[(1 - h(t_i))\, S(\tau_{\kati-1})]}^{1 - d_i}\\
    &= {h(t_i)}^{d_i} \, {[1 - h(t_i)]}^{1-d_i} \, S(\tau_{\kati - 1}) \\
    &= {h(t_i)}^{d_i}  \, {[1 - h(t_i)]}^{1-d_i} \, \prod_{j=1}^{\kati-1} [1 - h(\tau_j)].
\end{align*}
With this formulation, the mean negative log-likelihood in~\eqref{eq:loss_general} can be rewritten as
\begin{align}
    \label{eq:loss_hazard}
    \text{loss}
    &= - \frac{1}{n} \sum_{i=1}^n \left(d_i \log[h(t_i \mid \x_i)] +  (1 - d_i) \log[1 - h(t_i \mid \x_i)] + 
    \sum_{j = 1}^{\kati-1} \log[1 - h(\tau_j \mid \x_i)] \right) \nonumber\\
    &= - \frac{1}{n} \sum_{i=1}^n \sum_{j=1}^\kati \left(y_{ij} \log[h(\tau_{j} \mid \x_i)] +  (1 - y_{ij}) \log[1 - h(\tau_{j} \mid \x_i)] \right).
\end{align}
Here, $y_{ij} = \mathbbm{1}\{t_i = \tau_j,\, d_i = 1\}$, meaning $\mathbf y_i = (y_{i1}, \ldots , y_{im})$ is a vector of zeros with a single 1 at the event index $\kappa(t_i)$ when $t_i$ corresponds to an observed event ($d_i = 1$).
We recognize this as the negative log-likelihood for Bernoulli data, or binary cross-entropy, a useful discovery first noted by \citet{brown1975use}.

With the two loss functions~\eqref{eq:loss_general} and~\eqref{eq:loss_hazard}, we can now make survival models by parameterizing the PMF or the hazard function and minimizing the corresponding loss.
For classical statistical models, these approaches are equivalent and have been used to obtain maximum likelihood estimates for the parameters in the PMF/hazard function; see, e.g., \citet{tutz2016} for a review.
We will, however, not consider classical maximum likelihood estimates, but focus on the part of the literature that fit neural networks for the purpose of time-to-event prediction, in which case the two loss functions may give different results.

\subsection{Parameterization with Neural Networks}
\label{sub:parameterization_with_neural_networks}

In the previous section, we saw that the survival likelihood can be expressed in terms of the PMF or the hazard function.
In the following, we will describe how to use this to create survival methods by parameterizing the PMF or hazard with neural networks.
In theory, as both approaches minimize the same negative log-likelihood, the methods should yield similar results. 
But as neural networks are quite complex, this might not be the case in practice.

First, considering the hazard parametrization of the likelihood, let $\phi(\x) \in \R^m$ represent a neural network that takes the covariates $\x$ as input and gives $m$ outputs, each corresponding to a discrete time-point $\tau_j$, i.e., $\phi(\x) = {\{\phi_1(\x), \ldots, \phi_m(\x)\}}$.
As the discrete hazards are (conditional) probabilities, we require $h(\tau_j \mid \x) \in [0, 1]$.
This can be achieved by applying the logistic function (sigmoid function) to the neural network
\begin{align*}
    h(\tau_j \mid \x) = \frac{1}{1 + \exp[-\phi_j(\x)]}.
\end{align*}
We can estimate the hazard function by minimizing the loss~\eqref{eq:loss_hazard}, and survival estimates can be obtained from~\eqref{eq:surv_hazard}.
To the best of our knowledge, this method was first proposed by \citet{gensheimer2019}. However, if one considers $\phi_j(\x)$ an arbitrary parametric function of $\x$, the approach is well known in the survival literature and seems to have been first addressed by \citet{Cox1972} and \citet{brown1975use}; see also \citet{10.2307/270718}.
The book by \citet{tutz2016} gives a review of the approach.

The implementation we use in the experiments in Sections~\ref{sec:simulations} and~\ref{sec:experiments} differs slightly from that of \citet{gensheimer2019}, as it was found to be more numerically stable (see Appendix~\ref{app:implementation_details}).
In this paper, we will refer to the method as \textit{Logistic-Hazard}, as coined by \citet{brown1975use} (one can also find the term \text{Logistic Discrete Hazard} used in the statistical literature).
\citet{gensheimer2019}, on the other hand,  referred to it as \textit{Nnet-survival}, but to be better able to contrast this method to the other methods presented in this paper, we will instead use the more descriptive \textit{Logistic-Hazard}.

We can obtain a survival model by parameterizing the PMF in a similar manner to the Logistic-Hazard method.
As for the hazards, the PMF represents probabilities $f(\tau_j \mid \x) \in [0, 1]$, but, contrary to the conditional probabilities that define the hazard, we now require the PMF to sum to 1.
As we only observe event times in $\mathbb{T}_C$, we fulfill this requirement indirectly through the probability of surviving past $\tau_m$, i.e.,
\begin{align}
    \label{eq:pmf_par_equal}
    \sum_{k=1}^m f(\tau_k \mid \x) + S(\tau_m \mid x) = 1.
\end{align}
Now, again with $\phi(\x) \in \R^{m}$ denoting a neural network, the PMF can be expressed as
\begin{align}
    \label{eq:pmf_f}
    f(\tau_j \mid \x) = \frac{\exp[\phi_j(\x)]}{1 + \sum_{k=1}^{m} \exp[\phi_k(\x)]}, \quad\quad \text{for } j = 1, \ldots, m.
\end{align}
Note that~\eqref{eq:pmf_f} is equivalent to the softmax function with a fixed $\phi_{m+1}(\x) = 0$.
Alternatively, one could let $\phi_{m+1}(\x)$ vary freely, something that is quite common in machine learning, but we chose to follow the typical conventions in statistics.
By combining~\eqref{eq:survival_orig} and~\eqref{eq:pmf_par_equal}, we can express the survival function as
\begin{align}
    \label{eq:s_pmf_sum}
    S(\tau_j \mid \x) &=  \sum_{k=j+1}^{m}  f(\tau_k \mid \x) +  S(\tau_m \mid \x), \quad\quad \text{for } j=1,\ldots,m-1, \\
     S(\tau_m \mid \x) &= \frac{1}{1 + \sum_{k=1}^m \exp[\phi_k(\x)]}. \nonumber
\end{align}
Now, let $\sigma_j[\phi(\x)]$, for $j=1,\ldots,m+1$,  denote the softmax in~\eqref{eq:pmf_f}, meaning $\sigma_{m+1}[\phi(\x)] = S(\tau_m \mid \x)$. Notice the similarities to classification with $m+1$ classes, as we are essentially classifying whether the event is happening at either time $\tau_1, \ldots, \tau_m$ or later than $\tau_m$. However, due to censoring, the likelihood is \textit{not} the cross-entropy.
Instead, by inserting~\eqref{eq:pmf_f} and~\eqref{eq:s_pmf_sum} into~\eqref{eq:loss_general}, we get the mean negative log-likelihood
\begin{align}
    \label{eq:loss_pmf_sigma}
    \text{loss} 
    &= -\frac{1}{n} \sum_{i=1}^n \left( d_i \log [\sigma_\kati(\phi(\x_i)) ] + 
    (1-d_i) \log \left[\sum_{k=\kati+1}^{m+1} \sigma_k(\phi(\x)) \right] \right),
\end{align}
where $\kati$ still denotes the duration index of individual $i$'s event time, i.e., $t_i = \tau_\kati$.
This is essentially the same negative log-likelihood as presented by \citet{deephit}, but with only one type of event.
Also, note that contrary to the work by \citet{deephit} the negative log-likelihood in~\eqref{eq:loss_pmf_sigma} allows for survival past time $\tau_m$.
Some numerical improvements of the implementation are addressed in Appendix~\ref{app:implementation_details}.
We will refer to this method simply by \textit{PMF} as this term is unambiguously discrete, contrary to the term \textit{hazard} which is used both for discrete and continuous time.

\subsubsection{Multi-Task Logistic Regression}
\label{sub:multi_task_logistic_regression}

Multi-task logistic regression, proposed by \citet{MTLR}, provides a generalization of the binomial log-likelihood to jointly model the sequence of binary labels $Y_j = \mathbbm{1}\{\Ts \leq \tau_j\}$.
This means that $Y = (y_1, \ldots , y_m)$ is a sequence with zeros for every time $\tau_j$ up to the event time, followed by one's, e.g., $(0, \ldots, 0, 1, \ldots 1)$. 
Then 
\begin{align}
    \label{eq:mtlr_orig}
    \Pm(Y = (y_1, \dots, y_m) \mid \x) =
    \frac{\exp\left[\sum_{k=1}^m y_k \psi_k(\x)\right]}
    {1 + \sum_{k=1}^{m} \exp\left[\sum_{l=k}^m \psi_l(\x)\right]}.
\end{align}
\citet{MTLR} only consider linear predictors $\psi_k(\x) = \x^T \boldsymbol\beta_k$, but this was extended to a neural network by \citet{fotso2018}.
The parameters of $\psi_k(\x)$ are found by minimizing the negative log-likelihood in~\eqref{eq:loss_general}.

As $f(\tau_j \mid \x) = \Pm(Y=(y_1, \ldots, y_m) \mid \x)$, where $y_k = \mathbbm{1}\{k \geq j\}$, the expression in~\eqref{eq:mtlr_orig} can be written as
\begin{align*}
    f(\tau_j \mid \x) &= \frac{\exp\left[\sum_{k=j}^m \psi_k(\x)\right]} {1 + \sum_{k=1}^{m} \exp\left[\sum_{l=k}^m \psi_l(\x)\right]}
    = \frac{\exp[\phi_j(\x)]} {1 + \sum_{k=1}^{m} \exp[\phi_k(\x)]},
\end{align*}
where $\phi_j(\x) = \sum_{k=j}^m \psi_k(\x)$. Hence, the multi-task logistic regression is equivalent to the PMF in~\eqref{eq:pmf_f}, but where $\phi_j(\x)$ is the (reverse) cumulative sum of the output of the network $\psi(\x) \in \R^m$.
To the extent of our knowledge, there are no benefits to this extra cumulative sum. Instead, it simply requires unnecessary computations, and, for large $m$, it can cause numerical instabilities.
Hence, we will not consider this method further.

\section{Continuous-Time Models}
\label{sec:continuous_time_models}

In the following, we no longer consider the time scale to be discrete, but instead consider continuous-time models, where $T^*, C^* > 0$, and we let $T = \min\{T^*, C^*\}$ and $D = \mathbbm{1}\{\Ts \leq C^*\}$ be as before.
Let $\tau$ denote the maximum possible value of $C^*$, meaning $P(C^* \leq \tau) = 1$.
Hence, a potentially right-censored observation $T$ is in the interval $T \in \interv{0}{\tau}$.
Instead of a PMF, we now have the density function $f(t)$ and the continuous-time survival function
\begin{align*}
    S(t) = \Pm(T^* > t) = \int_t^\tau f(z)\, dz + S(\tau).
\end{align*}
Furthermore, the continuous-time hazard rate is a non-negative function of the time (no longer restricted to $[0, 1]$),
\begin{align}
    \label{eq:hazard_cont_def}
    h(t) = \frac{f(t)}{S(t)} = \lim_{\Delta t \rightarrow 0} \frac{\Pm(t \leq T^* < t + \Delta t \mid T^* \geq t)}{\Delta t}.
\end{align}
As a result, we can express the survival function in terms of the cumulative hazard $H(t) = \int_{\tau_0}^t h(z)\, dz$,
\begin{align}
    \label{eq:surv_continuous_cumulative_hazard}
    S(t) = \exp[-H(t)].
\end{align}
This yields the continuous-time version of the likelihood contribution in~\eqref{eq:likelihood_survival},
\begin{align}
    \label{eq:likelihood_cont}
    L_i = {f(t_i)}^{d_i}\, {S(t_i)}^{1-d_i} = {h(t_i)}^{d_i}\, S(t_i) = {h(t_i)}^{d_i}\, \exp[-H(t_i)].
\end{align}
The derivation of $L_i$ follows the same steps as the derivation of the discrete likelihood contribution~\eqref{eq:likelihood_survival}, only with density functions instead of probability mass functions.

In what follows, we will first discuss how we can apply the discrete-time methods from Section~\ref{sub:parameterization_with_neural_networks} for continuous-time data.
We will here address how time can be discretized to fit the discrete-time model formulation, and how to interpolate an estimated discrete survival function for continuous-time predictions.
Then, we will propose a new continuous-time method by assuming that the hazard in~\eqref{eq:hazard_cont_def} is piecewise constant, which we call \textit{PC-Hazard}.

\subsection{Discretization of Durations}
\label{sub:discretization_of_durations}

\begin{figure}[tpb]
    \centering
    \includegraphics[width=0.7\linewidth]{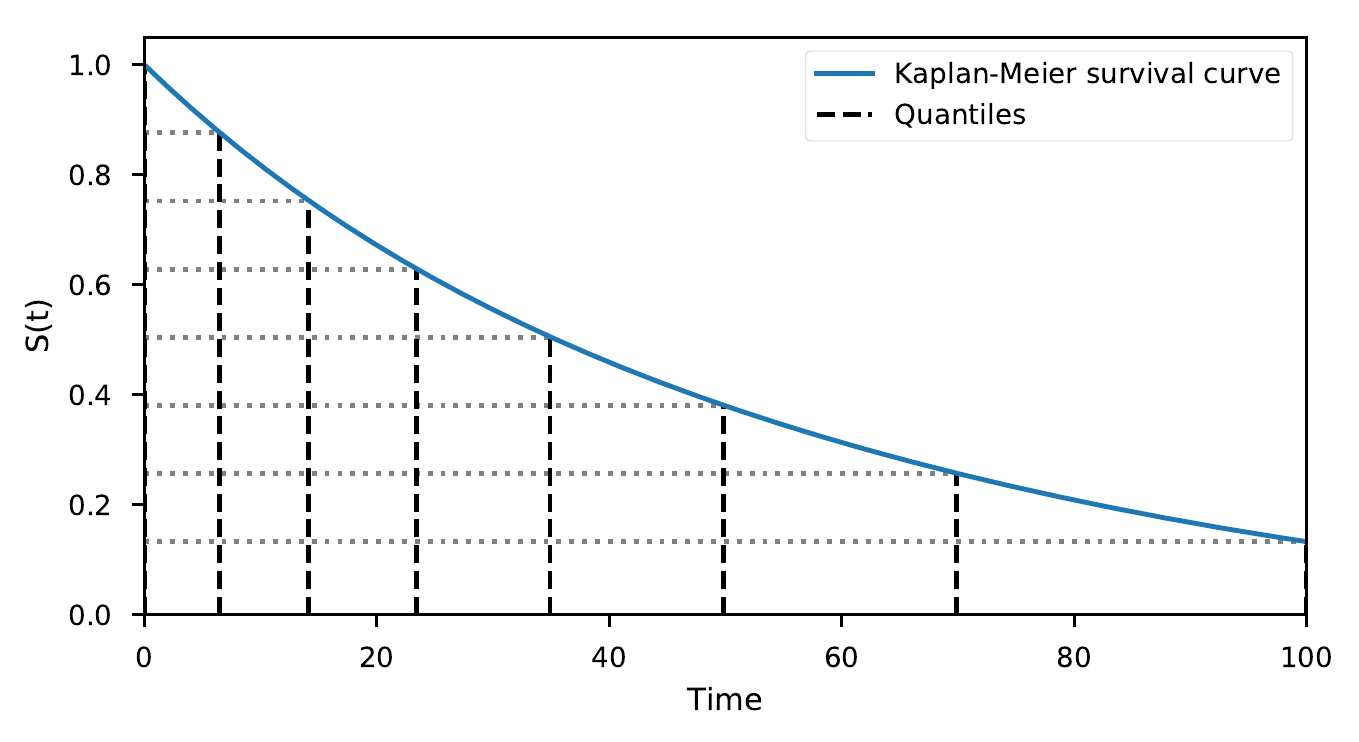}
    \vspace*{-3mm}
    \caption{Illustration of the Kaplan-Meier based discretization scheme. The quantiles of the Kaplan-Meier curve are used as the grid points.}\label{fig:km_discretization}
\end{figure}

Both the PMF and Logistic-Hazard methods require time to be discrete on the form $0 = \tau_0 < \tau_1 < \cdots < \tau_m$.
Hence, to apply the methods to continuous-time data, we need to perform some form of discretization of the time scale
(also, for inherently discrete event times, we might want to coarsen the discrete time scale to obtain a smaller subset of $\tau_j$'s, as this will reduce the number of parameters in the neural networks).
Possibly the most obvious way to discretize time is to make an equidistant grid in $[0, \tau]$ with $m$ grid points.
An alternative, that we explore in this paper, is to make a grid based on the distribution of the event times.
By estimating the survival function $S(t)$ with the Kaplan-Meier estimator, we obtain a general trend of event times.
With $\hat S(t)$ denoting the Kaplan-Meier survival estimates, we can make a grid from the quantiles of the estimates, $1 = \hat S(0) = \zeta_0 > \zeta_1 > \cdots > \zeta_m = \hat S(\tau)$.
We will assume that each interval has the same decrease in the survival estimate, so that $\zeta_j - \zeta_{j+1} = (1 - \hat S(\tau)) / m$. 
The corresponding duration grid, $\tau_1 < \cdots < \tau_m$, is found by solving $\hat S(\tau_j) = \zeta_j$.
We will then obtain a more dense grid in intervals with  more events, and less dense grid in intervals with fewer events.
This is illustrated in Figure~\ref{fig:km_discretization}, where we can see that the grid becomes coarser as the slope of the survival curve becomes less steep.

The discrete-time methods assume that all events and censorings occur at the $\tau_j$'s, so, when performing the discretization, we move all event times in an interval to the end of that interval while censored times are moved to the end of the previous interval.
This means that for $\tau_{j-1} < T_i \leq \tau_j$, we replace $T_i$ by $\tau_j$ if $D_i = 1$, and by $\tau_{j-1}$ if $D_i = 0$.
Our reason for this choice is that this is typically how event times are recorded. Consider a study where we are only able to make observations at times $\tau_1 < \tau_2 < \cdots < \tau_m$.
For a censored observation, $\tau_{j-1}$ is the last point in time where the individual was recorded alive, while for an observed event, $\tau_j$ is the first duration for which the individual was recorded with the event.

As a side note, an alternative way to obtain the discrete loss in~\eqref{eq:loss_hazard} is by assuming continuous event times in defined intervals $\intervl{\tau_j}{\tau_{j+1}}$ and censorings that only occur at the beginning or end of the intervals \citep[see, e.g.,][]{tutz2016}.
This justifies the use of this loss for continuous-time data grouped in intervals.

\subsection{Interpolation for Continuous-Time Predictions}
\label{sub:interpolation_for_continuous_time_predictions}

\begin{figure}[t]
    \centering
    \includegraphics[width=0.7\linewidth]{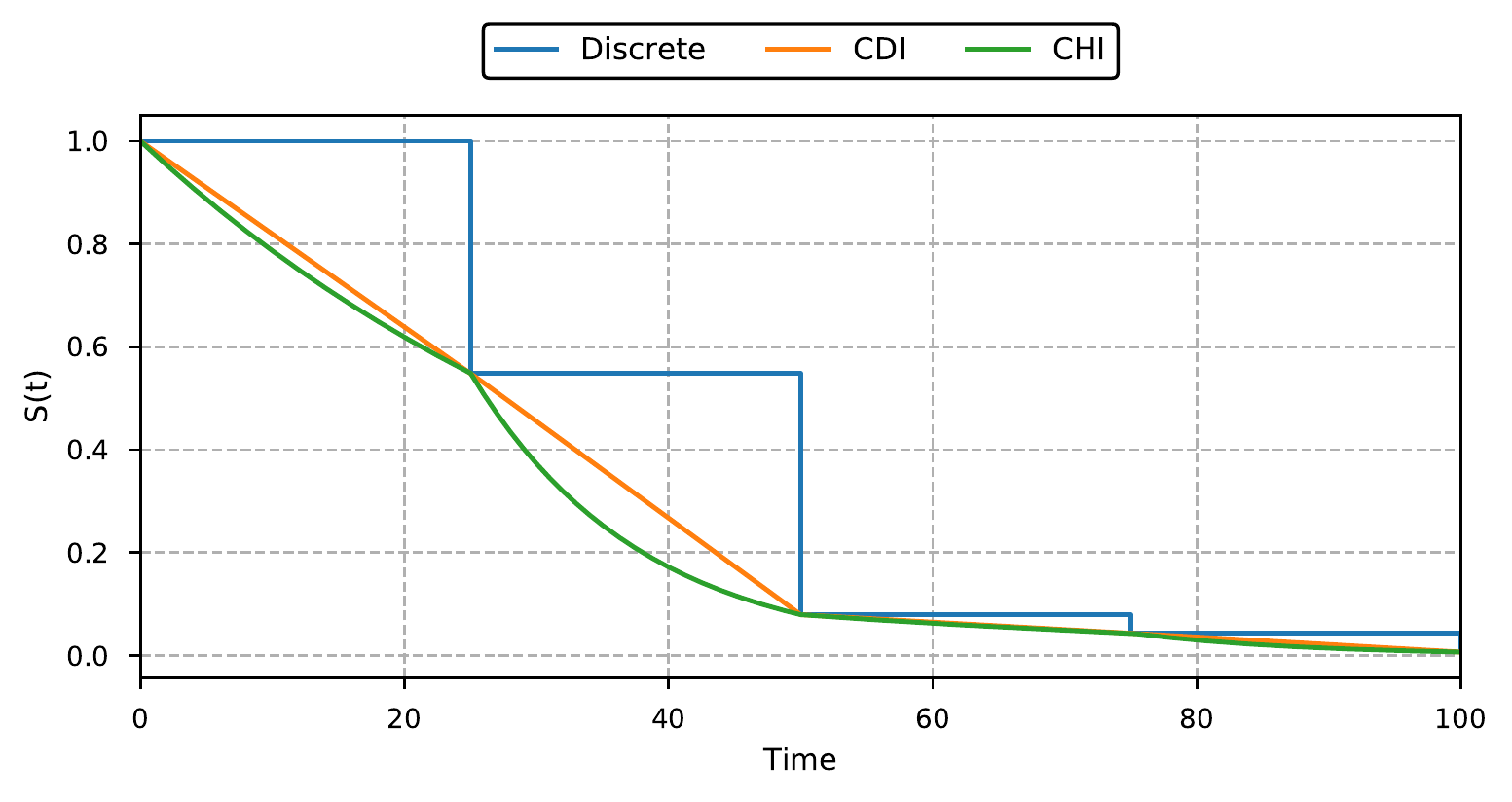}
    \vspace*{-3mm}
    \caption{Survival estimates by a discrete model (e.g., PMF or Logistic-Hazard) for 5 grid points. The three lines represent the discrete survival estimates and the two interpolation schemes in Section~\ref{sub:interpolation_for_continuous_time_predictions}: The constant density interpolation (CDI) and constant hazard interpolation (CHI).}\label{fig:hazard_interpolation}
\end{figure}

When discrete-time survival methods are applied to continuous-time data, as described in Section~\ref{sub:discretization_of_durations}, the survival estimates become a step function with steps at the grid points (blue line in Figure~\ref{fig:hazard_interpolation}).
Consequently, for coarser grids,  it might be beneficial to interpolate the discrete survival estimates.
In this regard, we propose two simple interpolation schemes that fulfill the monotonicity requirement of the survival function.
The first assumes that the probability density function is constant in each time interval $\interv{\tau_{j-1}}{\tau_{j}}$, while the second scheme assumes constant hazards in each time interval. This corresponds to piecewise linear and piecewise exponential survival estimates, respectively, and we will, therefore, refer to the schemes as \textit{constant density interpolation} (CDI) and \textit{constant hazard interpolation} (CHI).
See Figure~\ref{fig:hazard_interpolation} for an illustration of the two schemes and the discrete survival estimates.

\subsubsection{Constant Density Interpolation (CDI)}
\label{sub:constant_pdf_interpolation}

For a continuous time $t \in \interv{\tau_{j-1}}{\tau_j}$, linear interpolation of the discrete survival function takes the form
\begin{align*}
    S(t) = S(\tau_{j-1}) + \left[S(\tau_j)  - S(\tau_{j-1}) \right] \, \frac{t - \tau_{j-1}}{\Delta \tau_j},
\end{align*}
where $\Delta \tau_j = \tau_j - \tau_{j-1}$.
This means that the density function $f(t)$ is constant in this interval
\begin{align}
    f(t) = - S'(t) = \frac{S(\tau_{j-1}) - S(\tau_{j})}{\Delta \tau_j}.
\end{align}
Let us now rewrite the expression of the survival function as
\begin{align*}
    S(t) 
    &= \frac{S(\tau_{j-1}) \tau_j - S(\tau_j) \tau_{j-1}}{\Delta \tau_j} - \frac{S(\tau_{j-1}) - S(\tau_{j})}{\Delta \tau_j}\, t 
    = \alpha_j - \beta_j\, t,
\end{align*}
where both $\alpha_j$ and $\beta_j$ are non-negative.
Using that the density is $f(t) = -S'(t) = \beta_j$, we get a simple expression for the hazard function~\eqref{eq:hazard_cont_def}
\begin{align*}
    h(t) = \frac{f(t)}{S(t)} = \frac{\beta_j}{\alpha_j - \beta_j\, t}.
\end{align*}
Hence, we see that linear interpolation of the survival function corresponds to a constant density function and an increasing hazard rate throughout the interval.

\subsubsection{Constant Hazard Interpolation (CHI)}
\label{sub:constant_hazard_interpolation}

The following scheme  assumes constant hazard in each interval, which corresponds to linear interpolation of the cumulative hazard function.
For a continuous time $t \in \interv{\tau_{j-1}}{\tau_j}$, the interpolated cumulative hazard is then
\begin{align*}
    H(t) = H(\tau_{j-1}) + \left[H(\tau_j)  - H(\tau_{j-1}) \right] \, \frac{t - \tau_{j-1}}{\Delta \tau_j}.
\end{align*}
This means that the  hazard function is constant in this interval
\begin{align*}
    h(t) = H'(t) = \frac{H(\tau_{j}) - H(\tau_{j-1})}{\Delta \tau_j} = \eta_j,
\end{align*}
and from~\eqref{eq:surv_continuous_cumulative_hazard}, we obtain the piecewise exponential survival function
\begin{align*}
    S(t)
    = \exp[-H(t)]
    = \exp\left[- \eta_j\, (t - \tau_{j-1}) \right]\, S(\tau_{j-1}).
\end{align*}
Finally, the density is
\begin{align*}
    f(t) = h(t)\, S(t) =  \eta_j\, S(t),
\end{align*}
showing that the density is decreasing throughout the interval.

Summarizing the two interpolation methods, CDI assumes that the events are spread evenly in the interval, while the CHI assumes that there are more events at the beginning of the interval.
Correspondingly, the CDI assumes that the longer an individual ``survives'' in an interval, the higher the risk becomes of experiencing an event in the next immediate moment (increasing hazard), contrary to the CHI which assumes that this risk is constant.
In fact, in the next section, we will propose a new method with the same assumptions as CHI, but contrary to the CHI, we can train it on continuous-time data.

\subsection{A Piecewise Constant Continuous-Time Hazard Parametrization}
\label{sub:continuous_time_hazard_parametrization}

We now propose a continuous-time method by parameterizing the hazards in~\eqref{eq:likelihood_cont}.
As for the CHI in Section~\ref{sub:constant_hazard_interpolation}, we will let the continuous-time hazard be piecewise constant.
Disregarding the neural networks, this model was first proposed by
\citet{holford1976life}, and further developed by \citet{friedman1982piecewise} who found that piecewise constant hazards yields a likelihood proportional to that of a Poisson likelihood;
see Appendix~\ref{app:pc_hazard_as_poisson_regression} for details.

Consider a partition of the time scale $0 = \tau_0 < \tau_1 < \cdots < \tau_m = \tau$, and let $\kappa(t)$ denote the interval index of time $t$ such that $t \in \interv{\tau_{\kat-1}}{\tau_\kat}$ (this is slightly different from the discrete case where we had $t = \tau_\kat$).
If we assume that the hazard is constant within each interval, we can express the hazard as the step function
\begin{align*}
    h(t) = \eta_\kat,
\end{align*}
for a set of non-negative constants $\{\eta_1, \ldots, \eta_m\}$.
For $\Delta \tau_j = \tau_j - \tau_{j-1}$, we can now express the cumulative hazard as 
\begin{align*}
    H(t)
    &= \int_{0}^{t} h(z) dz\\
    &= \left(\sum_{j=1}^{\kat - 1} \int_{\tau_{j-1}}^{\tau_j} h(z) dz\right) + \int_{\kat -1}^{t} h(z) dz\\
    &= \left(\sum_{j=1}^{\kat - 1} \eta_j\, \Delta \tau_j\right) +  \eta_\kat\,  (t - \tau_{\kat -1 }).
\end{align*}
Inserting this into~\eqref{eq:likelihood_cont} yields  the likelihood contribution for individual $i$  
\begin{align}
    \label{eq:like_contrib_pc-hazard}
    L_i
    &= {h(t_i)}^{d_i}\, \exp\left[- H(t_i) \right]
    = {\eta_\kati}^{d_i}\, \exp\left[-\eta_\kati\, (t - \tau_{\kati - 1})\right]
    \prod_{j=1}^{\kati - 1} \exp\left[-\eta_j\, \Delta \tau_{j}\right].
\end{align}
What remains is to parameterize the hazard with a neural network.
However, to avoid passing all the $\tau_j$'s to the loss function, we let the network instead parameterize the quantities $\tilde \eta_j = \eta_j\, \Delta \tau_k$.
This allows us to rewrite the likelihood contribution as
\begin{align*}
    L_i
    &= {\left(\frac{{\tilde\eta_\kati{}}}{\Delta \tau_\kati}\right)}^{d_i}\, \exp\left[-\tilde\eta_\kati\, \rho(t_i) \right]
    \prod_{j=1}^{\kati - 1} \exp\left[-\tilde\eta_j\right] \\
    &\propto {\tilde\eta_\kati{}}^{d_i}\, \exp\left[-\tilde\eta_\kati\, \rho(t_i) \right]
    \prod_{j=1}^{\kati - 1} \exp\left[-\tilde\eta_j\right],
\end{align*}
where
\begin{align}
    \label{eq:rho_frac}
    \rho(t) = \frac{t - \tau_{\kat -1}}{\Delta \tau_\kat},
\end{align}
is the proportion of interval $\kat$ at time $t$.

As before, let $\phi(\x) \in \R^m$ denote a neural network.
To ensure that $\tilde{\eta}_j$ is non-negative, we use the softplus function
\begin{align}
    \label{eq:pc_hazard_softplus}
    \tilde{\eta}_j = \log(1 + \exp[\phi_j(\x)]).
\end{align}
Our model can now be fitted by  minimizing the mean negative log-likelihood
\begin{align*}
    \text{loss} &= - \frac{1}{n}\sum_{i=1}^n \left( d_i\, \log \tilde{\eta}_\kati (\x_i)  -  \tilde{\eta}_\kati (\x_i)\, \rho(t_i)  - \sum_{j=1}^{\kati - 1} \tilde{\eta}_j (\x_i) \right),
\end{align*}
and estimates for the survival function can be obtained by
\begin{align}
    \label{eq:cont_haz_surv}
    S(t \mid \x) = \exp[- H(t \mid \x)] = \exp[-\tilde{\eta}_\kat (\x)\, \rho(t)] \prod_{j=1}^{\kat-1} \exp[-\tilde{\eta}_j(\x)],
\end{align}
where $\rho(t)$ is given by~\eqref{eq:rho_frac}.
We will refer to this method as the \textit{piecewise constant hazard} method, or \textit{PC-Hazard}.
Even though this is a continuous-time method, we still need to decide the set of $\tau_j$'s that define the intervals.
Therefore, the discretization techniques discussed in Section~\ref{sub:discretization_of_durations} are also relevant for this method.

Comparing the PC-Hazard to the Logistic-Hazard method with survival estimates interpolated with CHI (Section~\ref{sub:constant_hazard_interpolation}), we see that the only difference is in the loss function, as both PC-Hazard and CHI have piecewise constant hazards.
In other words, the two methods both use~\eqref{eq:cont_haz_surv} to obtain survival estimates, but they have different estimates for the $\tilde \eta_j$'s as the PC-Hazard use the observed continuous event times and censoring times, while Logistic-Hazard discretizes the times to a predefined set of $\tau_j$'s as described in Section~\ref{sub:discretization_of_durations}.

\section{Simulations}
\label{sec:simulations}

To get a better understanding of the  methodologies discussed in Sections~\ref{sec:discrete_time_models} and~\ref{sec:continuous_time_models}, we perform a simulation study where we vary the size of the training sets, the discretization scheme, and the number of grid points used for discretization. 
\citet{gensheimer2019} performed a similar study to evaluate the effect of discretization on their Logistic-Hazard method with the conclusion that there were no differences in performance.  However, their simulations were quite simple (only one binary covariate), their only performance metric was the \citet{Harrell1982} concordance at 1-year survival, and they did not include any interpolation of the survival estimates.
For this reason, we find that further investigation is warranted.

\begin{figure}[tpb]
    \centering
    \includegraphics[width=1\linewidth]{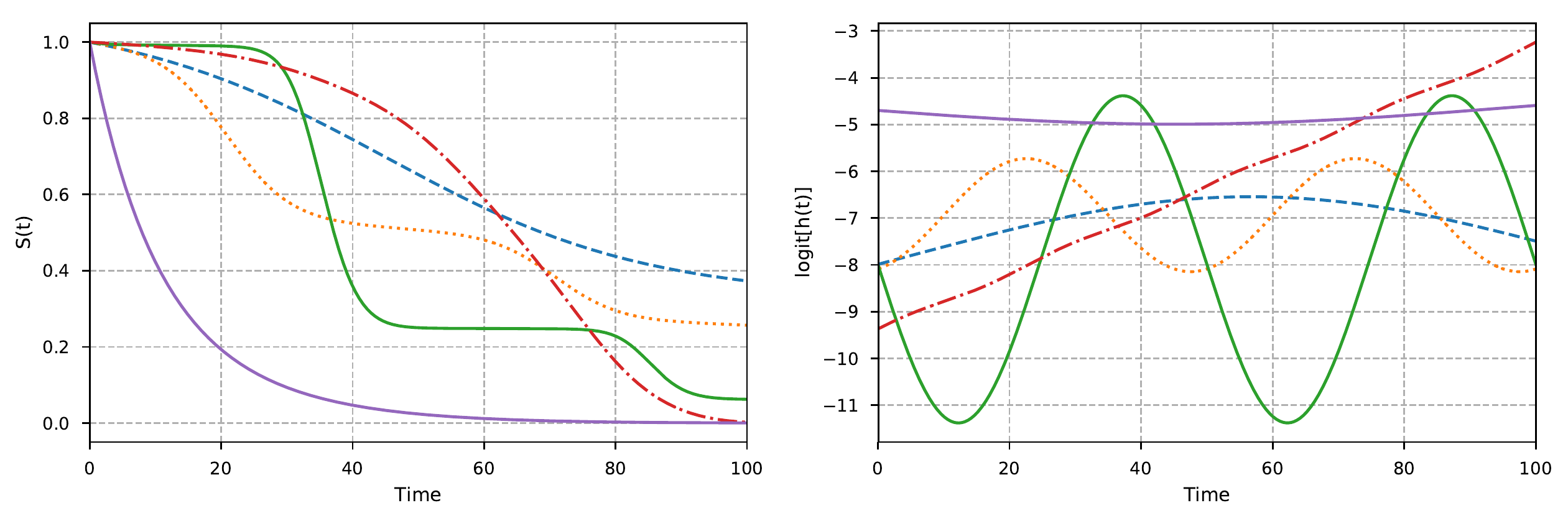}
    \vspace*{-6mm}
    \caption{Examples from the simulation study in Section~\ref{sec:simulations}. The left figure shows examples of 5 simulated survival curves, while the right figure shows the corresponding logit hazards.}\label{fig:sim_examples}
\end{figure}

We generate simulated survival times by sequentially sampling from discrete-time hazards defined on a fine grid of time points.
The hazards are specified through their logit transforms, as this enables us to use functions in $\mathbb R$ while still obtaining hazards in $[0, 1]$.
The logit hazards, $g(t) = \text{logit}[h(t)]$, are linear combinations of the three functions
\begin{align*}
    &g_\text{sin}(t \mid \x)  = \gamma_1 \sin(\gamma_2 [ t + \gamma_3]) + \gamma_4,\\
    &g_\text{con}(t \mid \x)  = \gamma_5, \\
    &g_\text{acc}(t \mid \x)  = \gamma_6 \cdot t - 10,
\end{align*}
with additional parameters $\gamma_7$, $\gamma_8$, and $\gamma_9$ determining the linear combination.
As described in Appendix~\ref{app:simulation}, each $\gamma_k$ is a function of five covariates, meaning we have a total of 45 covariates.
We let the discrete time scale consist of 1,000 equidistant points between 0 and 100 (i.e.\ $\tau_0=0$, $\tau_1 = 0.1$, $\tau_2 = 0.2$, \ldots, $\tau_{1000} = 100$).
Knowing the hazards, the true survival function can be obtained with~\eqref{eq:surv_hazard}, $S(\tau_j \mid \x) = \prod_{k=1}^j [1 - h(\tau_k \mid \x)]$. 
In Figure~\ref{fig:sim_examples} we show five examples of logit hazard functions and their corresponding survival functions.
Note that even though we simulate our data using a discrete-time model, the time-grid is so fine that this mimics simulation from a continuous-time model.
The full details of this simulation study are given in Appendix~\ref{app:simulation}.

We created three training sets of size 3,000, 10,000, and 50,000, a validation set of size 10,000 (for hyperparameter tuning) and a test set of size 100,000.
For the training and validation sets, we included a censoring distribution with constant hazard resulting in 37 \% censoring. The full uncensored test set is used for evaluation.
For the discretization of the time scale,  we applied both the equidistant scheme and the Kaplan-Meier quantiles, each with 5, 25, 100, and 250 grid points. 
The neural networks were all ReLU-nets with batch normalization and dropout between each layer, with all layers consisting of the same number of nodes.
We performed a hyperparameter grid search over 1, 2, 4, and 8 layers; 16, 64, and 256 nodes; and dropout of 0 and 0.5.
Each net was trained with batch size of 256 and the AdamWR optimizer~\citep{adamwr} with cycle length 1, where, at each restart, the cycle length was doubled and  the learning rate was multiplied by 0.8. 
Learning rates were found using the methods proposed by \citet{smith2017}.
The hyperparameter tuning was repeated 10 times, giving 10 fitted models for each combination of method, grid size, discretization scheme, and training set size.

\subsection{Comparison of Discrete-Time Methods}
\label{sub:comparison_of_discrete_methods}

We start by comparing the two discrete methods from Section~\ref{sub:parameterization_with_neural_networks}, that parameterize the PMF and the discrete-time hazards. We refer to them as PMF and  Logistic-Hazard, respectively.
For evaluation, we use the time-dependent concordance \citep{Antolini2005}, in addition to
the MSE between the survival estimates and the true survival function at all 1,000 time points $\tau_1, \ldots, \tau_{1000}$
\begin{align}
    \label{eq:mse_surv}
    \text{MSE} = \frac{1}{100,000} \sum_{i=1}^{100,000} \frac{1}{1,000}\sum_{j=1}^{1,000} {\left(\hat S(\tau_j \mid \x_i) -  S_i(\tau_j) \right)}^2.
\end{align}
Here $\hat S(\tau_j \mid \x_i)$ and $S_i(\tau_j)$ are the estimated and true survival functions, respectively, for individual $i$ (in the test set) at time $\tau_j$.
So, in this regard, the discrete-time survival estimates are represented by step functions, as illustrated in Figure~\ref{fig:hazard_interpolation}.

\begin{figure}[tb]
    \centering
    \includegraphics[width=0.99\linewidth]{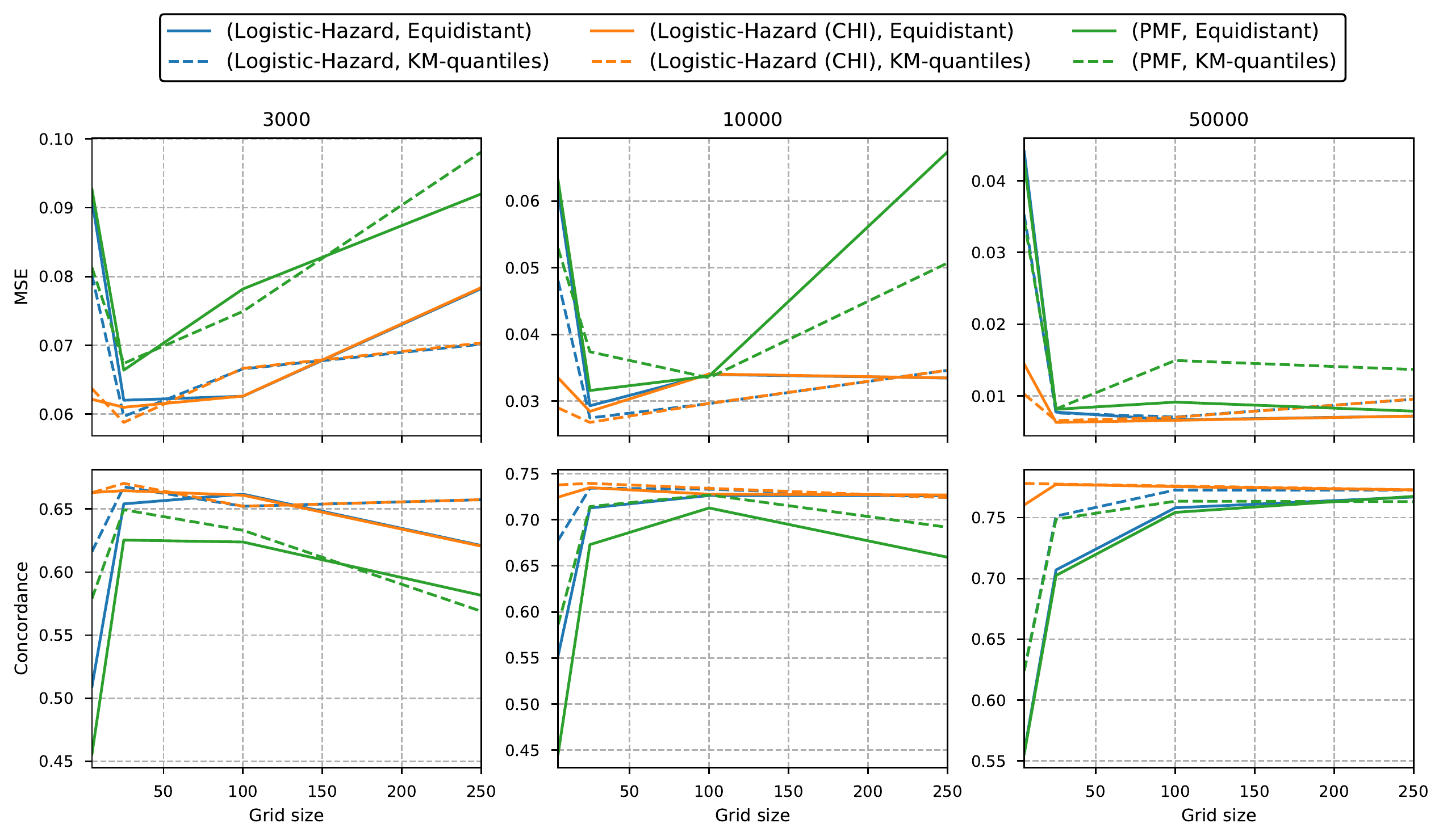}
    \vspace*{-1mm}
    \caption{Median MSE and concordance for each grid size in the simulation study in Section~\ref{sub:comparison_of_discrete_methods}. The number above each plot gives the size of the training set. The full lines use an equidistant grid, while the dotted lines use Kaplan-Meier quantiles for discretization. Note that the plots are not on the same scale.}\label{fig:sim_grid_haz_pmf_chi}
\end{figure}

In Figure~\ref{fig:sim_grid_haz_pmf_chi} we plot the median test scores of the two methods versus the grid size used for discretization. The numbers above each plot give the size of the training set used to fit the methods.
The full lines represent equidistant grids, while the dotted lines are from grids obtained with quantiles from Kaplan-Meier survival estimates.
We have also included the constant hazard interpolation (CHI) of the survival estimates from the Logistic-Hazard method (see Section~\ref{sub:constant_hazard_interpolation}).

It is evident that smaller discretization grids are better for the smaller training sets, while larger training sets allow for larger grids. This is reasonable as the smaller grids result in fewer parameters in the neural networks. Nevertheless, the smallest grid of size 5 seems to only work well for the interpolated estimates, and very poorly for the discrete estimates.
We also note that the discretization grids from Kaplan-Meier quantiles seem to give slightly better scores than the equidistant grids.
Comparing the discrete survival estimates from Logistic-Hazard (blue lines) with the CHI estimates (orange lines), we see that the two lines overlap for larger grid sizes. This is expected as the effect of interpolation decreases as the grids become denser.

In general, the PMF method does not perform as well as the Logistic-Hazard, though the difference is rather small.
Also, while the interpolated estimates yield better results for most grid configurations, the best scores are almost identical.
This means that the interpolated estimates have more stable performance, but with careful tuning of the discretization scheme, similar performance can be obtained with the discrete estimates.

\subsection{Comparison of Interpolation Schemes}
\label{sub:comparison_of_interpolation_schemes}

\begin{figure}[tb]
    \centering
    \includegraphics[width=0.99\linewidth]{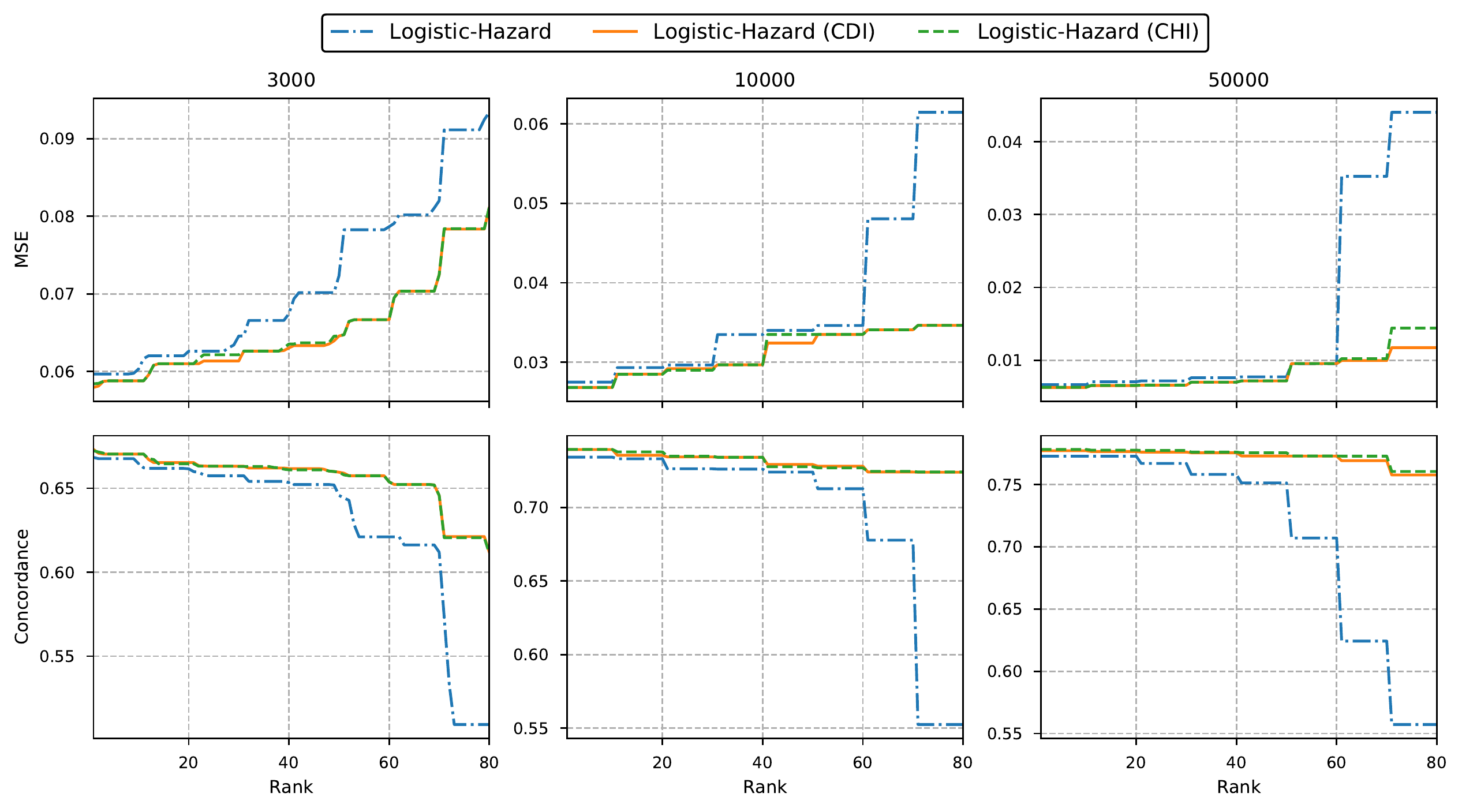}
    \vspace*{-1mm}
    \caption{MSE and concordance from the simulation study in Section~\ref{sub:comparison_of_interpolation_schemes}. The scores are plotted from best to worst. The number above each plot gives the size of the training set. Note that the plots are not on the same scale.}\label{fig:sim_rank_hazard_ip}
\end{figure}

In the following, we compare the interpolation schemes for the discrete-time hazard method Logistic-Hazard. The experiments are not shown for the PMF method as the results are very similar.

In Section~\ref{sub:interpolation_for_continuous_time_predictions} we presented two methods for interpolation of discrete survival estimates. The first assume constant density in each interval (denoted CDI for constant density interpolation), while the second assumes constant hazard in each interval (denoted CHI for constant hazard interpolation).
In our simulation study, we have four grid sizes and two discretization schemes. As the hyperparameter tuning was repeated 10 times this gives 80 fitted models for each method on each data set. 
In Figure~\ref{fig:sim_rank_hazard_ip}, we plot the scores of these 80 models sorted from best to worst, as this both tells us the best performance, in addition to the stability of the methods.
The figure contains results from the discrete survival estimates (Logistic-Hazard), the constant density interpolation (CDI), and the constant hazard interpolation (CHI).

Clearly, there is almost no difference in performance between the two interpolation schemes, while the discrete estimates have slightly worse best-case performance and much worse worst-case performance.
In fact, the only difference between the two interpolation schemes is that that CDI estimates give slightly better MSE while the CHI estimates give slightly better concordance.
In this regard, we will in the further simulations only include the CHI estimates, as they have the same assumption as the continuous-time PC-Hazard method, simplifying the comparison between the methods.

\subsection{Comparison with PC-Hazard}
\label{sub:comparison_with_the_continuous_hazard_method}

\begin{figure}[tb]
    \centering
    \includegraphics[width=0.99\linewidth]{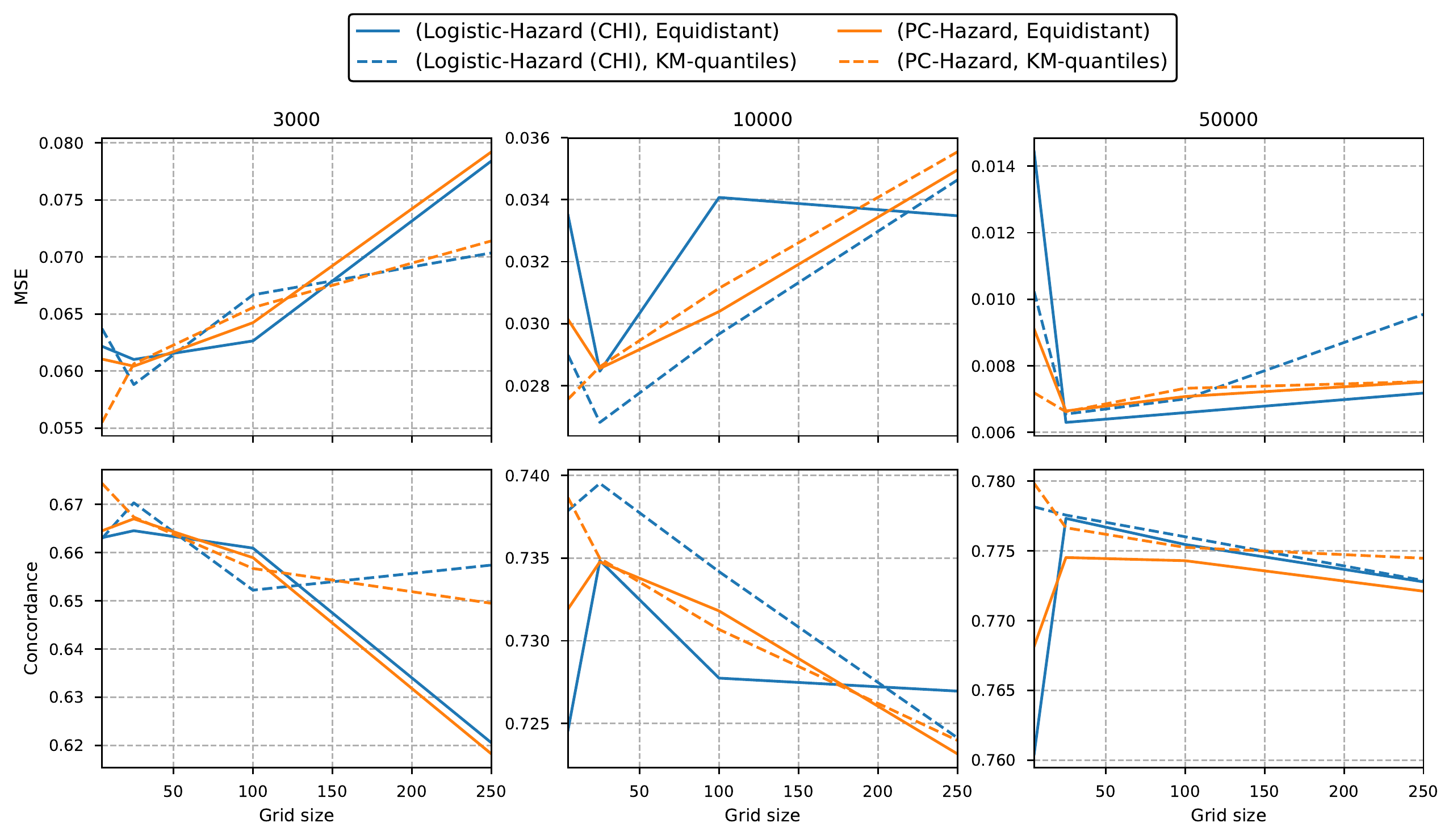}
    \caption{Median MSE and concordance for each grid size of the simulation study in Section~\ref{sub:comparison_with_the_continuous_hazard_method}. The number above each plot gives the size of the training set. The full lines use an equidistant grid, while the dotted lines use Kaplan-Meier quantiles for discretization. Note that the plots are not on the same scale.}\label{fig:sim_grid_cont_chi}
    \vspace*{-1mm}
\end{figure}

Finally, we compare the previous methods with our proposed continuous-time hazard method 
from Section~\ref{sub:continuous_time_hazard_parametrization}, PC-Hazard.
In Figure~\ref{fig:sim_grid_cont_chi} we plot the MSE and concordance for the interpolated Logistic-Hazard (CHI) method, and the continuous-time PC-Hazard method.
First, we notice that PC-Hazard does better for the smallest grids with only five grid points, while Logistic-Hazard (CHI) typically performs best with 25 grid points.
Also, in terms of MSE, PC-Hazard does the best for the smallest training set, while Logistic-Hazard (CHI) does better for the two larger training sets.
In terms of concordance, PC-Hazard performs the best for the smallest and largest data sets.
All differences are however quite small.
On the other hand, the Logistic-Hazard (CHI) estimates do better for a variety of grid configurations, showing that it is less sensitive to the discretization than the PC-Hazard method.
Finally, we again see that the Kaplan-Meier quantiles seem to give slightly better performance than the equidistant discretization grids.

In Figure~\ref{fig:sim_rank_all} in Appendix~\ref{app:simulation}, we have included a plot of the same type as Figure~\ref{fig:sim_rank_hazard_ip} for the Logistic-Hazard (CHI) method, the Logistic-Hazard method, the PMF method and the PC-Hazard method.
The figure again shows that the PMF method performs slightly worse than the other methods, while the PC-Hazard method performs similarly to the Logistic-Hazard (CHI) estimates.

\subsection{Summary of Simulations}
\label{sub:summary_of_simulations}

To summarize the results of the simulations, we have shown that the size of the discretization grid (number of $\tau_j$'s) has a large impact on the performance of the methods, and therefore needs to be carefully tuned.
Finer grids enable the methods to reduce bias in the predictions but require more parameters in the neural networks (higher variance).
By defining the discretization grid with Kaplan-Meier quantiles, the performance for the smaller grids typically improve.
Furthermore, interpolation of the discrete-time survival estimates made the performance less sensitive to the number of grid points, and was generally found to improve performance of the methods for smaller grid sizes.
The performance of the two proposed interpolation schemes, CHI and CDI, was more or less indistinguishable.

Comparing the three methods, we found that PMF performs slightly worse than Logistic-Hazard, both in terms of best-case performance and stability to discretization-grid configurations. PC-Hazard was found to be competitive with the interpolated Logistic-Hazard method and even performed better for the smallest training set.
But the differences between all methods were small, and the size of the training sets and the grid size were shown to have a much larger impact on the performance than the choice of method.

\section{Experiments with Real Data}
\label{sec:experiments}

We now compare the methods discussed in this paper to other methods in the literature, in particular DeepHit~\citep{deephit}, DeepSurv~\citep{DeepSurv}, Cox-Time~\citep{Cox-Time}, CoxCC~\citep{Cox-Time}, Random Survival Forests~\citep[RSF,][]{Ishwaran2008}, and a regular Cox regression.

We conduct the comparison on five common real-world data sets:
the Study to Understand Prognoses Preferences Outcomes and Risks of Treatment (SUPPORT), the Molecular Taxonomy of Breast Cancer International Consortium (METABRIC), the Rotterdam tumor bank and German Breast Cancer Study Group (Rot. \& GBSG),
 the Assay Of Serum Free Light Chain (FLCHAIN),
 and the National Wilm's Tumor Study (NWTCO).
 \citet{DeepSurv} made the first three datasets available in their python package DeepSurv, and we have made no further preprocessing of the data.
FLCHAIN and NWTCO were made available in the survival package of R \citep{survival-package}, but we use the same version of FLCHAIN as~\citet{Cox-Time}. No alterations were made to the NWTCO data set.
The size, the number of covariates, and the proportion of censored individuals in each data set are given in Table~\ref{tab:datasets}.

\begin{table}[tb]
    \centering
    \begin{tabular}{lrrr}
        \toprule
        Dataset &    Size &  Covariates &  Prop. Censored \\
        \midrule
        FLCHAIN          &  6,524 &         8 &   0.70 \\
        METABRIC         &  1,904 &         9 &   0.42 \\
        NWTCO            &  4,028 &         6 &   0.86 \\
        Rot. \& GBSG     &  2,232 &         7 &   0.43 \\
        SUPPORT          &  8,873 &        14 &   0.32 \\
        \bottomrule
    \end{tabular}
    \caption{Datasets for comparing survival methods. }\label{tab:datasets}
\end{table}
 
Hyperparameter tuning is performed with the evaluation criteria given in the paper that proposed the method.
For the methods presented in this paper, however, we will use the integrated Brier score (IBS) by~\citet{Graf1999} computed over 100 equidistant points between the minimum and maximum observed times in the validation set.
In the simulations in Section~\ref{sec:simulations}, we could use the validation loss for this purpose.
This is, however, no longer feasible as we now also need to tune the discretization scheme and the discretization affects the magnitude of the losses.
The IBS considers both discrimination and calibration, and is a useful substitute for the MSE~\eqref{eq:mse_surv} when the true survival function is not available. Hence, we believe it is a reasonable tuning criterion.

The experiments were conducted by five-fold cross-validation.
We used the same hyperparameter search and training strategy as presented in Section 6.1 of the paper by \citet{Cox-Time}, but decrease the learning rate by 0.8 at the start of each cycle, as this was found to give more stable training.
The best parameter configuration for each method on each fold was fitted 10 times, and we calculated the median concordance and integrated Brier score (IBS) of the 10 repetitions and averaged that over the five folds. The results are presented in Tables~\ref{tab:concordance_all} and~\ref{tab:ibs_all}.

In terms of concordance, we see that DeepHit and PC-Hazard perform very well. 
The three Logistic-Hazard methods and Cox-Time all perform close to the PC-Hazard, while the PMF, RSF and the other Cox methods perform slightly worse.
The concordances of the two proposed interpolation schemes, CHI and CDI, are very similar, but the CDI method gives slightly higher scores.  
There does, however, not seem to be much performance gain in interpolation for the concordance.

\begin{table}[tb]
    \resizebox{\textwidth}{!}{%
        \centering
        \begin{tabular}{lrrrrr}
            \toprule
            Model &         FLCHAIN &        METABRIC &           NWTCO &    Rot. \& GBSG &         SUPPORT \\
            \midrule
            Cox Regression         &           0.790 &           0.626 &           0.706 &           0.664 &           0.599 \\
            CoxCC                  &           0.792 &           0.647 &           0.711 &           0.670 &           0.614 \\
            DeepSurv               &           0.792 &           0.640 &           0.709 &           0.674 &           0.615 \\
            Cox-Time               &  \textbf{0.793} &           0.664 &           0.709 &           0.674 &           0.630 \\
            RSF                    &           0.784 &           0.651 &           0.705 &           0.668 &           0.632 \\
            DeepHit                &           0.791 &  \textbf{0.675} &           0.710 &           0.675 &  \textbf{0.639} \\
            PMF                    &           0.786 &           0.632 &           0.710 &           0.669 &           0.627 \\
            Logistic-Hazard        &           0.792 &           0.658 &           0.704 &           0.670 &           0.625 \\
            Logistic-Hazard (CHI)  &           0.790 &           0.656 &           0.714 &           0.673 &           0.628 \\
            Logistic-Hazard (CDI)  &           0.790 &           0.660 &           0.700 &           0.676 &           0.630 \\
            PC-Hazard              &           0.791 &           0.655 &  \textbf{0.716} &  \textbf{0.679} &           0.628 \\
            \bottomrule
        \end{tabular}
    }
    \caption{Concordance from 5-fold cross-validation on real-world data sets.}\label{tab:concordance_all}
\end{table}

\begin{table}[tb]
    \resizebox{\textwidth}{!}{%
        \centering
        \begin{tabular}{lrrrrr}
            \toprule
            Model &          FLCHAIN &         METABRIC &            NWTCO &     Rot. \& GBSG &          SUPPORT \\
            \midrule
            Cox Regression         &           0.0961 &           0.183 &           0.0791 &           0.180 &           0.218 \\
            CoxCC                  &           0.0924 &           0.173 &           0.0745 &           0.171 &           0.213 \\
            DeepSurv               &           0.0919 &           0.175 &           0.0745 &           0.170 &           0.213 \\
            Cox-Time               &           0.0925 &           0.173 &           0.0753 &           0.170 &   \textbf{0.212} \\
            RSF                    &           0.0928 &           0.175 &           0.0749 &           0.170 &           0.213 \\
            DeepHit                &           0.0929 &           0.186 &           0.0758 &           0.184 &           0.227 \\
            PMF                    &           0.0924 &           0.174 &           0.0748 &  \textbf{0.169} &           0.213 \\
            Logistic-Hazard        &           0.0918 &  \textbf{0.172} &           0.0742 &           0.171 &           0.213 \\
            Logistic-Hazard (CHI)  &           0.0919 &           0.173 &  \textbf{0.0738} &           0.170 &           0.213 \\
            Logistic-Hazard (CDI)  &  \textbf{0.0917} &  \textbf{0.172} &           0.0741 &           0.170 &   \textbf{0.212} \\
            PC-Hazard              &           0.0918 &  \textbf{0.172} &  \textbf{0.0738} &  \textbf{0.169} &   \textbf{0.212} \\
            \bottomrule
        \end{tabular}
    }
    \caption{Integrated Brier score from 5-fold cross-validation on real-world data sets.}\label{tab:ibs_all}
\end{table}

Examining the IBS in Table~\ref{tab:ibs_all} we again find that PC-Hazard performs very well.
But now, DeepHit does quite poorly. This is expected as DeepHit is designed for discrimination rather than well-calibrated estimates~\citep[see][]{Cox-Time}.
In general, the PMF, the RSF and the three proportional Cox methods seem to have slightly higher IBS than the Hazard methods, but again the differences are quite small.
Cox-Time performs quite well on all data sets except for FLCHAIN and NWTCO\@.
Comparing the interpolation schemes of Logistic-Hazard, it seems that CDI still performs slightly better than CHI, although both are quite close to the discrete estimates of Logistic-Hazard.

Interestingly, we see that, for the NWTCO data set, PC-Hazard and Logistic-Hazard (CHI) performs the best both in terms of concordance and IBS\@.
This likely means that the piecewise exponential survival estimates are a good way of representing this data set.

In summary, all three methods discussed in this paper are competitive with existing survival methodology. However, the interpolated Logistic-Hazard or the PC-Hazard seems to give the most stable high performance considering both discrimination and calibration.

\section{Discussion}
\label{sec:discussion}

In this paper, we have explored survival methodology built on neural networks for discrete-time data, and how it can be applied for continuous-time prediction.
We have compared two existing discrete-time survival methods that minimize the negative log-likelihood of right-censored event times, where the first method~\citep{deephit} parameterize the event-time probability mass function (PMF), while the second method~\citep{gensheimer2019} parameterize the discrete hazard rate (Logistic-Hazard).
Furthermore, we showed that the multi-task logistic regression \citep{MTLR, fotso2018} is, in fact, a PMF parametrization.
Through empirical studies of simulated and real data sets, we found that the Logistic-Hazard method performed slightly better than the PMF parametrization.

We proposed two interpolation schemes for the discrete methods, which were found to typically improve performance for smaller data sets.
This is likely caused by the fact that interpolation allows for coarser discretization of the time scale, which reduces the number of parameters in the neural networks.
We found that the interpolation scheme that assumed constant density within each time interval (CDI) performed slightly better than the scheme assuming constant hazard in each time interval (CHI). Note, however, that none of the schemes affect the training procedure, meaning both can be compared at test time.

We also proposed a new continuous-time method that assumes constant hazard in predefined time-intervals (PC-Hazard).
The method was found to perform very well compared to existing methods, both in terms of discrimination and calibration.
Furthermore, in a simulation study, we found that the method continued to performed well for coarser discretization grids than the interpolated Logistic-Hazard method. This was particularly beneficial for the smallest data set in the simulation study.

All three methods investigated in this paper need some form of discretization or coarsening of the time-scale.
In that regard, we proposed a simple scheme that use the quantiles of the event-time distribution estimated by Kaplan-Meier, and showed through simulations that the quantile-based grids typically outperformed equidistant grids for coarser grids.


\acks{This work was supported by The Norwegian Research Council 237718 through the Big Insight Center for research-driven innovation.}



\appendix
\setcounter{equation}{0}
\setcounter{table}{0}
\setcounter{figure}{0}
\renewcommand{\theequation}{\thesection.\arabic{equation}}
\renewcommand{\theHequation}{\thesection.\arabic{equation}}
\renewcommand{\thetable}{\thesection.\arabic{table}}
\renewcommand{\theHtable}{\thesection.\arabic{table}}
\renewcommand{\thefigure}{\thesection.\arabic{figure}}
\renewcommand{\theHfigure}{\thesection.\arabic{figure}}
\renewcommand{\theHsection}{\thesection}
\setcounter{equation}{0}
\setcounter{table}{0}
\setcounter{figure}{0}
\section{More on the Simulations}
\label{app:simulation}

In the following, we include additional information about the simulation study in Section~\ref{sec:simulations}.
We start by explaining in detail how the data sets were created, and in Section~\ref{app:additional_simulation_results} we give some additional results.

\subsection{Discrete-Time Survival Simulations from Logit Hazards}
\label{app:sim_details}

The simulated survival data sets were generated by drawing from the discrete hazard $h(t \mid \x)$ across times $t \in \{0.1, 0.2, \ldots, 100\}$.
The discrete hazard was defined through the logit hazard $g(t \mid \x) \in \mathbb{R}$,
\begin{align*}
    h(t \mid \x) = \frac{1}{1 + \exp[- g(t \mid \x)]},
\end{align*}
ensuring that $h(t \mid \x) \in (0,\, 1)$.
We let the logit hazard be a weighted sum of three different functions, 
$g_\text{sin}(t \mid \x)$, $g_\text{con}(t \mid \x)$,  and $g_\text{acc}(t \mid \x)$, giving
\begin{align*}
    g(t \mid \x) &= \alpha_1\, g_\text{sin}(t \mid \x) + \alpha_2\, g_\text{con}(t \mid \x) + \alpha_3\, g_\text{acc}(t \mid \x), \\
    g_\text{sin}(t \mid \x)  &= \gamma_1 \sin(\gamma_2 [ t + \gamma_3]) + \gamma_4,\\
    g_\text{con}(t \mid \x)  &= \gamma_5, \\
    g_\text{acc}(t \mid \x)  &= \gamma_6 \cdot t - 10, \\
    \alpha_i &= \frac{\exp(\gamma_{i+6})}{\sum_{j=1}^3 \exp(\gamma_{j+6})}, \quad \text{for } i=1, 2, 3.
\end{align*}
Here, we actually have covariate-dependent $\gamma_i$'s, i.e., $\gamma_i(\x)$, but we have omitted the $\x$ for readability.
Let $\tilde x_j$ be a linear combination of a subset of the covariates, $\tilde x_j = \x_j^T \boldsymbol\beta_j$ for $j=1, \dots, 9$, where the subsets are non-overlapping and of equal size (if the subsets are of size $m$, we have $\x \in \mathbb{R}^{9 m}$).
The $\gamma$'s in the study are defined as
\begin{align*}
    \gamma_1(\x) &= 5 \tilde x_1, \\
    \gamma_2(\x) &= \frac{2\,\pi}{100} \cdot 2^{\lfloor \frac{5}{2}(\tilde x_2 + 1) - 1 \rfloor},\\
    \gamma_3(\x) &= 15 \tilde x_3, \\
    \gamma_4(\x) &= 2 \tilde x_4 - 6 - |\gamma_1(\x)|, \\
    \gamma_5(\x) &= \frac{5}{2} (\tilde x_5 + 1) - 8, \\
    \gamma_6(\x) &= \frac{1}{1 + \exp[- \frac{6}{2} (\tilde x_6 + 1) + 5]}, \\
    \gamma_7(\x) &= 5 (\tilde x_7 + 0.6),\\
    \gamma_8(\x) &= 5 \tilde x_8,\\
    \gamma_{9}(\x) &= 5 \tilde x_{9},
\end{align*}
where $\lfloor z \rfloor$ is the floor operation.
We draw $\tilde x_j \overset{iid}{\sim} \text{Unif}[-1, \, 1]$, and $\beta_k \overset{iid}{\sim} N(0,\, 1)$.  
The forms of the $\gamma_i(\x)$'s have been chosen to obtain reasonable survival functions. In particular, $\gamma_2(\x)$ ensures that the number of periods is a multiple of 2, as we found it more reasonable than having arbitrary periods.

Finally, we draw covariates $x_{j,k}$, while ensuring $\x_j^T \boldsymbol\beta_j = \tilde x_j$, through the following scheme:
For known $\boldsymbol\beta_j \in \mathbb{R}^m$, we draw $x_{j,k}$ conditionally such that
\begin{align*}
    \left(\tilde x_j - \sum_{i=1}^{k} x_{j, i}\, \beta_{j, i} \right) \mid \tilde x_j, x_{j,1}, \ldots, x_{j, k-1} \sim \text{Unif}[-1,\, 1], \quad \text{for } k=1,\ldots,m-1.
\end{align*}
Hence, we sample $u_{j,k} \overset{iid}{\sim} \text{Unif}[-1,\, 1]$ for $k=1, \ldots , m-1$, and set
$u_{j,k} = \tilde x_j - \sum_{i=1}^{k} x_{j, i}\, \beta_{j, i}$, giving the covariates
\begin{align*}
    x_{j,k} =
    \begin{cases}
        \frac{1}{\beta_{j,1}} \left(\tilde x_j - u_{j, 1}\right), & \text{if } k = 1\\
        \frac{1}{\beta_{j, k}}\left(u_{j, k-1} - u_{j, k}\right), & \text{if } k=2, \ldots, m-1\\
        \frac{1}{\beta_{j, m}} \left( \tilde x_j - \sum_{i=1}^{m-1} x_{j, i} \beta_{j, i} \right), & \text{if } k = m.
    \end{cases}
\end{align*}
Using this scheme, it is straightforward to change the number of covariates without affecting the hazards.
The code for generating theses simulations is available at \url{https://github.com/havakv/pycox}.

\subsection{Additional Simulation Results}
\label{app:additional_simulation_results}

\begin{figure}[tb]
    \centering
    \includegraphics[width=0.99\linewidth]{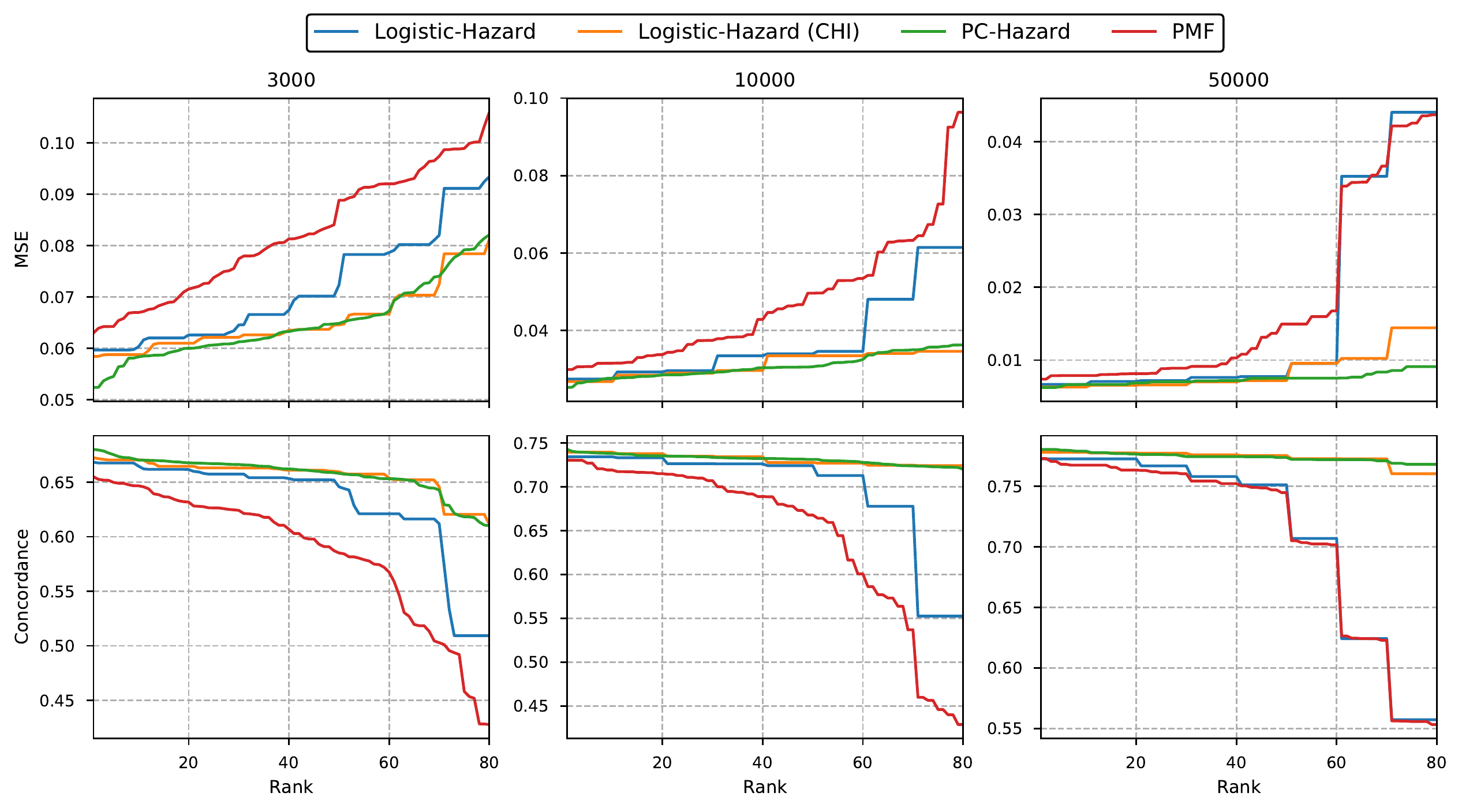}
    \vspace*{-1mm}
    \caption{MSE and concordance from the simulation study in Section~\ref{sec:simulations}. The scores are plotted from best to worst. The number above each plot gives the size of the training set. Note that the plots are not on the same scale.}\label{fig:sim_rank_all}
\end{figure}

We here present some additional results from the simulation study in Section~\ref{sub:comparison_of_interpolation_schemes}.
Recall that  each method is fitted 80 times (4 grids $\times$ 2 discretization schemes $\times$ 10 repetitions). 
In the same manner as in Figure~\ref{fig:sim_rank_hazard_ip}, we plot in Figure~\ref{fig:sim_rank_all} the MSE and concordance for the Logistic-Hazard, Logistic-Hazard (CHI), PC-Hazard, and PMF, where the scores of the 80 models are sorted from best to worst.

We again see that PC-Hazard and the Logistic-Hazard (CHI) perform better than the discrete estimates of Logistic-Hazard and PMF\@.
Furthermore, Logistic-Hazard seems to generally perform better than the PMF method.
We still find that for the best grid configurations, the differences between all models are very small.
But we reiterate that, for practical purposes, it is quite desirable to have stable performance for a variety of hyperparameter configurations.

\setcounter{equation}{0}
\setcounter{table}{0}
\setcounter{figure}{0}
\section{PC-Hazard and Poisson Regression}
\label{app:pc_hazard_as_poisson_regression}

The PC-Hazard method presented in Section~\ref{sub:continuous_time_hazard_parametrization} is essentially a neural network version of the piecewise exponential model studied by \citet{holford1976life} and \citet{friedman1982piecewise}.
Friedman showed that the likelihood obtained with piecewise constant hazards is proportional to the Poisson likelihood.
Consequently, one can use standard software to fit the model. 
Nevertheless, we prefer to implement the log-likelihood of PC-Hazard more directly, and do not use the Poisson likelihood.
This is because we wanted to ensure numerical stability with the softplus~\eqref{eq:pc_hazard_softplus} (as the inverse link function), while the Poisson likelihood available in most frameworks requires the log link function (i.e., an exponential activation function instead of the softplus).

To see how we can obtain the Poisson likelihood, we first need to define some variables.
Recall that $\kat$ denotes the index of an interval, such that $t \in \interv{\tau_{\kat-1}}{\tau_\kat}$.
If we define $y_{ij} = \mathbbm{1}\{\kati = j,\, d_i = 1\}$ and let
\begin{align*}
    \Delta\tilde{t}_{ij}=\left\{
        \begin{array}{ll}
            \tau_j - \tau_{j-1}, &\text{if } t_i > \tau_j\\
            t_i - \tau_{j-1},  &\text{if } \tau_{j-1} < t_i \leq \tau_j\\
            0, &\text{if } t_i \leq \tau_{j-1},
        \end{array}
        \right.
\end{align*}
we can rewrite the likelihood contribution in~\eqref{eq:like_contrib_pc-hazard} as
\begin{align*}
    L_i &=  \prod_{j=1}^{\kati} {(\Delta\tilde{t}_{ij} \eta_j)}^{y_{ij}}\, \exp \left[- \Delta\tilde{t}_{ij} \eta_j \right],
\end{align*}
which is proportional to the likelihood of $\kati$ independent Poisson-distributed variables $y_{ij}$ with expectations $\mu_{ij} = \Delta\tilde{t}_{ij} \eta_j$.

\setcounter{equation}{0}
\setcounter{table}{0}
\setcounter{figure}{0}
\section{Implementation details}
\label{app:implementation_details}

The implementations of the survival methods described in Sections~\ref{sec:discrete_time_models} and~\ref{sec:continuous_time_models} are slightly different from the mathematical notation.
This is because we also need to consider numerical stability. 
An implementation of the methods can be found at \url{https://github.com/havakv/pycox}.

For the PMF parameterization, we used the log-sum-exp trick 
\begin{align*}
    \log\left(\sum_j \exp(z_j) \right) = \gamma + \log\left(\sum_j \exp(z_j-\gamma) \right), 
\end{align*}
where $\gamma = \max_j(z_j)$, to ensure that we only take the exponential of non-positive numbers.
Hence, by rewriting the loss~\eqref{eq:loss_pmf_sigma} in terms of $\phi_j(\x)$, with $\phi_{m+1}(\x) = 0$ and $\gamma_i = \max_j (\phi_j(\x_i))$, we obtain
\begin{align*}
    \text{loss} = &- \frac{1}{n} \sum_{i=1}^n d_i [\phi_\kati (\x_i) - \gamma_i] +
        \frac{1}{n}\sum_{i=1}^n \log \left( \sum_{j=1}^{m+1} \exp[\phi_j(\x_i) - \gamma_i]\right) \\
        &- \frac{1}{n} \sum_{i=1}^n (1-d_i) \log \left(  \sum_{j= \kati+1}^{m+1} \exp[\phi_j(\x_i)-\gamma_i] \right).
\end{align*}

For the discrete hazard parametrization, we can simply formulate it as the negative log-likelihood for Bernoulli data, or binary cross-entropy, and use existing implementations of the loss function to ensure numerical stability. In practice, these implementations use the log-sum-exp trick on the logits $\phi_j(\x)$.

Finally, for the continuous hazard parametrization, we use existing implementations of the softplus function which uses a linear function over a certain threshold, meaning $\log(1 + \exp[z]) \approx z$ for large values of $z$.
However, we also note that for $z \approx 0$, we have that $\log(1 + z) \approx z$. Hence, for $\phi_\kati(\x_i) \ll 0$ we use that
\begin{align*}
    \log \tilde{\eta}_\kati(\x_i) = \log[\log(1 + \exp[\phi_\kati(\x_i)])] \approx \phi_\kati(\x_i).
\end{align*}



\bibliography{bibliography}

\begin{thebibliography}{26}
\providecommand{\natexlab}[1]{#1}
\providecommand{\url}[1]{\texttt{#1}}
\expandafter\ifx\csname urlstyle\endcsname\relax
  \providecommand{\doi}[1]{doi: #1}\else
  \providecommand{\doi}{doi: \begingroup \urlstyle{rm}\Url}\fi

\bibitem[Allison(1982)]{10.2307/270718}
Paul~D. Allison.
\newblock Discrete-time methods for the analysis of event histories.
\newblock \emph{Sociological Methodology}, 13:\penalty0 61--98, 1982.

\bibitem[Antolini et~al.(2005)Antolini, Boracchi, and Biganzoli]{Antolini2005}
Laura Antolini, Patrizia Boracchi, and Elia Biganzoli.
\newblock A time-dependent discrimination index for survival data.
\newblock \emph{Statistics in Medicine}, 24\penalty0 (24):\penalty0 3927--3944,
  2005.

\bibitem[Brown(1975)]{brown1975use}
Charles~C. Brown.
\newblock On the use of indicator variables for studying the time-dependence of
  parameters in a response-time model.
\newblock \emph{Biometrics}, 31\penalty0 (4):\penalty0 863--872, 1975.

\bibitem[Ching et~al.(2018)Ching, Zhu, and Garmire]{ching2018cox}
Travers Ching, Xun Zhu, and Lana~X. Garmire.
\newblock Cox-nnet: An artificial neural network method for prognosis
  prediction of high-throughput omics data.
\newblock \emph{PLoS Computational Biology}, 14\penalty0 (4):\penalty0
  e1006076, 2018.

\bibitem[Cox(1972)]{Cox1972}
David~R. Cox.
\newblock Regression models and life-tables.
\newblock \emph{Journal of the Royal Statistical Society. Series B
  (Methodological)}, 34\penalty0 (2):\penalty0 187--220, 1972.

\bibitem[Faraggi and Simon(1995)]{Faraggi1995}
David Faraggi and Richard Simon.
\newblock A neural network model for survival data.
\newblock \emph{Statistics in Medicine}, 14\penalty0 (1):\penalty0 73--82,
  1995.

\bibitem[Fotso(2018)]{fotso2018}
Stephane Fotso.
\newblock Deep neural networks for survival analysis based on a multi-task
  framework.
\newblock \emph{arXiv preprint arXiv:1801.05512}, 2018.

\bibitem[Friedman(1982)]{friedman1982piecewise}
Michael Friedman.
\newblock Piecewise exponential models for survival data with covariates.
\newblock \emph{The Annals of Statistics}, 10\penalty0 (1):\penalty0 101--113,
  1982.

\bibitem[Gensheimer and Narasimhan(2019)]{gensheimer2019}
Michael~F. Gensheimer and Balasubramanian Narasimhan.
\newblock A scalable discrete-time survival model for neural networks.
\newblock \emph{PeerJ}, 7:\penalty0 e6257, 2019.

\bibitem[Graf et~al.(1999)Graf, Schmoor, Sauerbrei, and Schumacher]{Graf1999}
Erika Graf, Claudia Schmoor, Willi Sauerbrei, and Martin Schumacher.
\newblock Assessment and comparison of prognostic classification schemes for
  survival data.
\newblock \emph{Statistics in Medicine}, 18\penalty0 (17-18):\penalty0
  2529--2545, 1999.

\bibitem[Harrell~Jr et~al.(1982)Harrell~Jr, Califf, Pryor, Lee, and
  Rosati]{Harrell1982}
Frank~E. Harrell~Jr, Robert~M. Califf, David~B. Pryor, Kerry~L. Lee, and
  Robert~A. Rosati.
\newblock Evaluating the yield of medical tests.
\newblock \emph{Journal of the American Medical Association}, 247\penalty0
  (18):\penalty0 2543--2546, 1982.

\bibitem[Holford(1976)]{holford1976life}
Theodore~R. Holford.
\newblock Life tables with concomitant information.
\newblock \emph{Biometrics}, pages 587--597, 1976.

\bibitem[Ishwaran et~al.(2008)Ishwaran, Kogalur, Blackstone, and
  Lauer]{Ishwaran2008}
Hemant Ishwaran, Udaya~B. Kogalur, Eugene~H. Blackstone, and Michael~S. Lauer.
\newblock Random survival forests.
\newblock \emph{Annals of Applied Statistics}, 2\penalty0 (3):\penalty0
  841--860, 2008.

\bibitem[Katzman et~al.(2018)Katzman, Shaham, Cloninger, Bates, Jiang, and
  Kluger]{DeepSurv}
Jared~L. Katzman, Uri Shaham, Alexander Cloninger, Jonathan Bates, Tingting
  Jiang, and Yuval Kluger.
\newblock Deepsurv: personalized treatment recommender system using a {C}ox
  proportional hazards deep neural network.
\newblock \emph{BMC Medical Research Methodology}, 18\penalty0 (1), 2018.

\bibitem[Klein and Moeschberger(2003)]{klein2005survival}
John~P. Klein and Melvin~L. Moeschberger.
\newblock \emph{Survival Analysis: Techniques for Censored and Truncated Data}.
\newblock Springer, New York, 2. edition, 2003.

\bibitem[Kvamme et~al.(2019)Kvamme, {{\O}}rnulf Borgan, and Scheel]{Cox-Time}
H{{\aa}}vard Kvamme, {{\O}}rnulf Borgan, and Ida Scheel.
\newblock Time-to-event prediction with neural networks and {C}ox regression.
\newblock \emph{Journal of Machine Learning Research}, 20\penalty0
  (129):\penalty0 1--30, 2019.

\bibitem[Lee et~al.(2018)Lee, Zame, Yoon, and van~der Schaar]{deephit}
Changhee Lee, William~R Zame, Jinsung Yoon, and Mihaela van~der Schaar.
\newblock Deephit: A deep learning approach to survival analysis with competing
  risks.
\newblock In \emph{Thirty-Second AAAI Conference on Artificial Intelligence},
  2018.

\bibitem[Loshchilov and Hutter(2019)]{adamwr}
Ilya Loshchilov and Frank Hutter.
\newblock Decoupled weight decay regularization.
\newblock In \emph{International Conference on Learning Representations}, 2019.

\bibitem[Luck et~al.(2017)Luck, Sylvain, Cardinal, Lodi, and Bengio]{Luck2017}
Margaux Luck, Tristan Sylvain, H{\'e}lo{\"\i}se Cardinal, Andrea Lodi, and
  Yoshua Bengio.
\newblock Deep learning for patient-specific kidney graft survival analysis.
\newblock \emph{arXiv preprint arXiv:1705.10245}, 2017.

\bibitem[{Smith}(2017)]{smith2017}
Leslie~N. {Smith}.
\newblock Cyclical learning rates for training neural networks.
\newblock In \emph{2017 IEEE Winter Conference on Applications of Computer
  Vision (WACV)}, pages 464--472, 2017.

\bibitem[Therneau(2015)]{survival-package}
Terry~M. Therneau.
\newblock \emph{A Package for Survival Analysis in S}, 2015.
\newblock URL \url{https://CRAN.R-project.org/package=survival}.
\newblock version 2.38.

\bibitem[Tutz and Schmid(2016)]{tutz2016}
Gerhard Tutz and Matthias Schmid.
\newblock \emph{Modeling discrete time-to-event data}.
\newblock Springer, 2016.

\bibitem[Yousefi et~al.(2017)Yousefi, Amrollahi, Amgad, Dong, Lewis, Song,
  Gutman, Halani, Vega, Brat, et~al.]{yousefi2017predicting}
Safoora Yousefi, Fatemeh Amrollahi, Mohamed Amgad, Chengliang Dong, Joshua~E.
  Lewis, Congzheng Song, David~A. Gutman, Sameer~H. Halani, Jose
  Enrique~Velazquez Vega, Daniel~J. Brat, et~al.
\newblock Predicting clinical outcomes from large scale cancer genomic profiles
  with deep survival models.
\newblock \emph{Scientific reports}, 7\penalty0 (11707), 2017.

\bibitem[Yu et~al.(2011)Yu, Greiner, Lin, and Baracos]{MTLR}
Chun-Nam Yu, Russell Greiner, Hsiu-Chin Lin, and Vickie Baracos.
\newblock Learning patient-specific cancer survival distributions as a sequence
  of dependent regressors.
\newblock In \emph{Advances in Neural Information Processing Systems 24}, pages
  1845--1853. Curran Associates, Inc., 2011.

\bibitem[{Zhu} et~al.(2017){Zhu}, {Yao}, {Zhu}, and {Huang}]{8100208}
X.~{Zhu}, J.~{Yao}, F.~{Zhu}, and J.~{Huang}.
\newblock Wsisa: Making survival prediction from whole slide histopathological
  images.
\newblock In \emph{2017 IEEE Conference on Computer Vision and Pattern
  Recognition (CVPR)}, pages 6855--6863, July 2017.

\bibitem[Zhu et~al.(2016)Zhu, Yao, and Huang]{7822579}
Xinliang Zhu, Jiawen Yao, and Junzhou Huang.
\newblock Deep convolutional neural network for survival analysis with
  pathological images.
\newblock In \emph{2016 IEEE International Conference on Bioinformatics and
  Biomedicine (BIBM)}, pages 544--547, 2016.

\end{thebibliography}

\end{document}